\documentclass[review,12pt,3p]{elsarticle}

\usepackage{framed,multirow}

\usepackage{amssymb}
\usepackage{amsmath}
\usepackage{latexsym}
\usepackage{soul}
\usepackage{url}
\usepackage{xcolor}
\definecolor{newcolor}{rgb}{.8,.349,.1}

\usepackage{lineno}
\usepackage{hyperref}
\usepackage{color}

\def \eg{\textit{e.g.}}
\def \ie{\textit{i.e.}}

\newcommand{\myparagraph}[1]{\noindent \textbf{#1}}

\journal{[Accepted for publication] J. Vis. Com. and Image Rep.}

\begin{document}

 

\begin{frontmatter}

\title{3D human pose estimation from depth maps using a deep combination of poses}


\author[label1,label2]{Manuel J. Mar\'in-Jim\'enez\corref{cor1}}
\address[label1]{Departamento de Inform\'atica y An\'alisis Num\'erico, Campus de Rabanales, Universidad de C\'ordoba, 14071, C\'ordoba, Spain}
\address[label2]{Instituto Maim\'onides de Investigaci\'on en Biomedicina (IMIBIC). Avenida Men\'endez Pidal s/n, 14004, C\'ordoba, Spain}
\cortext[cor1]{Corresponding author}
\ead{mjmarin@uco.es}
\ead[url]{http://www.uco.es/investiga/grupos/ava/}

\author[label1]{Francisco J. Romero-Ramirez}
\ead{fj.romero@uco.es}

\author[label1,label2]{Rafael Mu\~noz-Salinas}
\ead{rmsalinas@uco.es}

\author[label1,label2]{Rafael Medina-Carnicer}
\ead{rmedina@uco.es}

\begin{abstract}
Many real-world applications require the estimation of human body joints for higher-level tasks as, for example, human behaviour understanding. In recent years, depth sensors have become a popular approach to obtain three-dimensional information. The depth maps generated by these sensors provide information that can be employed to disambiguate the poses observed in two-dimensional images. This work addresses the problem of 3D human pose estimation from depth maps employing a Deep Learning approach. We propose a model, named \textit{Deep Depth Pose} (DDP), which receives a depth map containing a person and a set of predefined 3D prototype poses  and returns the 3D position of the body joints of the person. In particular, DDP is defined as a ConvNet that computes the specific weights needed to linearly combine the prototypes for the given input.
We have thoroughly evaluated DDP on the challenging `ITOP' and `UBC3V' datasets, which respectively depict  realistic and synthetic samples, defining a new state-of-the-art on them.
\end{abstract}


\end{frontmatter}


\section{Introduction} \label{sec:intro}
Many real-world applications require the estimation of human body joints for higher-level tasks, \eg, human behaviour understanding \citep{Ferrari09,zhu2014eccv}, medical physical therapies \citep{obdrvzalek2012real,achilles2016miccai} or human-computer interaction \citep{huo2009amigas,Shotton2013pami}. This problem is known in the literature as \textit{human pose estimation} (HPE). The goal of HPE is to estimate the position and orientation of body limbs in single images or video sequences. Note that a body limb (\eg, lower-arm) is usually defined by two joints (\eg, wrist and elbow). So, detecting body joints is equivalent to estimate the pose of their respective body limbs.

Although great effort has been put into solving this problem in previous years, it is far from solved. The main challenge associated with this task is that human body is highly deformable and suffers from self-occlusions (\ie~one body limb may occlude partially or completely other). In addition, the vast variety of people clothing and camera viewpoints make this problem even more difficult.

HPE has been classically addressed by using either single RGB images~\citep{Agarwal06,Eichner2012ijcv,Cherian14,rogez2017lcr} or multiple cameras \citep{Shen2011,LopezQuintero2017}.
In recent years, depth sensors, like the Microsoft Kinect device, have become affordable and, therefore, popular. These devices provide for each image point its distance to the sensor, \ie~its \textit{depth}. Depth can be used as a rough estimation of the 3D position of the image points, thus, it can help to disambiguate relative positions between body parts. 

\begin{figure}[t]
\centering
  \includegraphics[width=0.95 \textwidth]{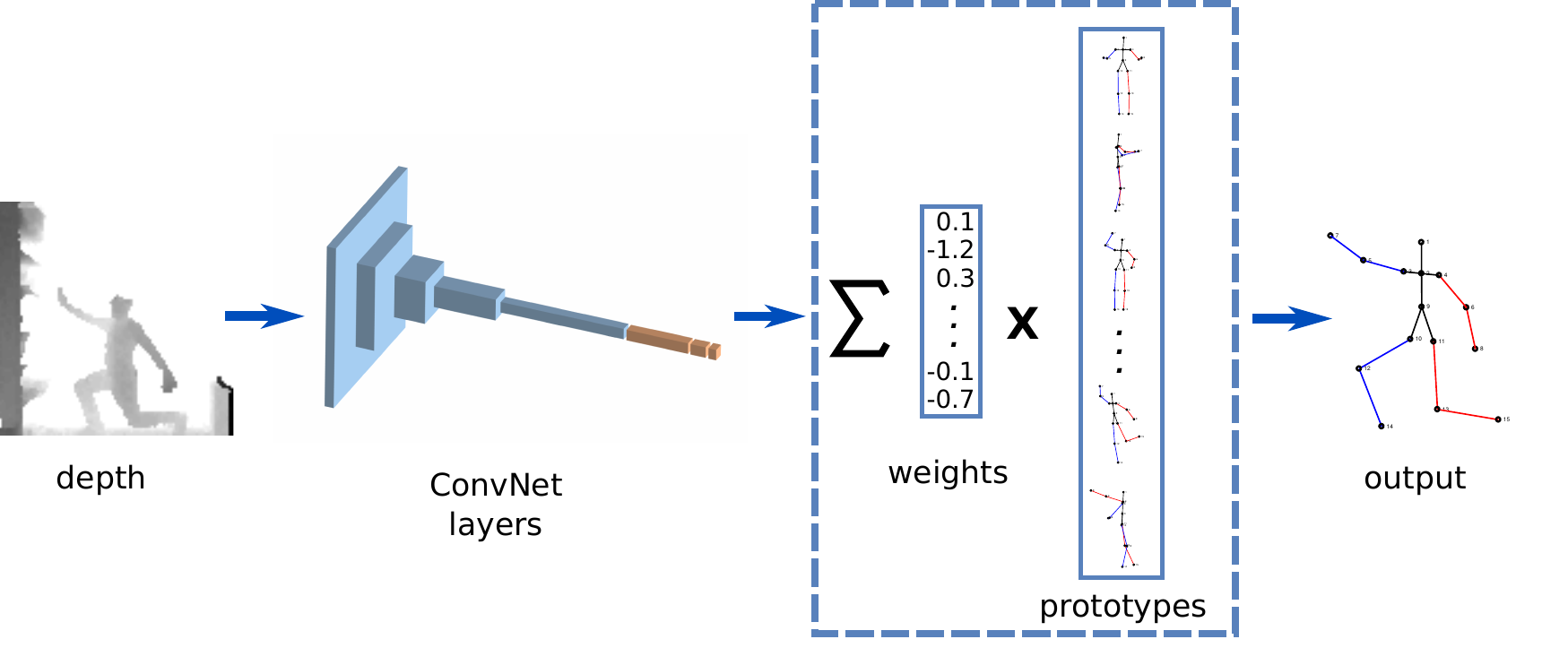}
  \caption{\textbf{DDP approach}. Given an input depth map, the 3D body pose is estimated as a linear combination of prototype poses, where the ConvNet is in charge of computing the weight of each prototype.}
  \label{fig:teaser}
\end{figure}

In this work, we propose the use of depth maps for 3D human pose estimation, using either single or multiple cameras. 
Our model, named \textit{Deep Depth Pose} (DDP), receives as input a depth map containing a person 
and a set of predefined 3D prototype poses and returns the 3D position of the body joints of the person. In particular, DDP computes the specific weights needed to linearly combine the prototypes for the given input (see Fig.~\ref{fig:teaser}). 
DDP is defined as a Convolutional Neural Network (ConvNet) \citep{Le1998} that computes the specific weights needed to linearly combine the prototypes for the given input depth map. For that purpose, a suitable architecture and loss function have been defined.
If multiple camera viewpoints are available, our system is able to fuse their data in order to provide a more accurate estimation of the body joint locations.


We have thoroughly evaluated our model on `ITOP' \cite{haque2016eccv} and `UBC3V' datasets \citep{Shafaei16}, establishing a new state-of-the-art on both datasets. For `ITOP' of $100\%$ of `\textit{accuracy at 10cm}' for both frontal and top views, versus the previous $77.4\%$ and $75.5\%$ reported by the authors of the dataset.
And, a $98.8\%$ of `\textit{accuracy at 10cm}' in `UBC3V' versus the previous $88.7\%$ reported by the authors of the dataset.

The rest of this paper is organized as follows. After presenting the related work, the proposed model is described in Sec.~\ref{sec:model}. Then, in Sec.~\ref{sec:dataset} the dataset used for evaluating our model is described. The experimental results are presented in Sec.~\ref{sec:expers}. Finally, Sec.~\ref{sec:conclus} concludes the paper.

\section{Related work} \label{sec:relworks}
We start reviewing in this section those relevant HPE approaches that use either depth or disparity maps. Then, some selected works that use RGB images are commented. 
For further information on the topic, a good survey can be found in \citep{liu2015}.

One of the most popular approaches for HPE from depth maps is the one from \citep{Shotton11cvpr}, extended in \citep{Shotton2013pami}. They use Random Forests to classify each pixel into body parts. 3D joint locations are derived from the  labelled depth map by using a local mode-finding approach based on Mean Shift.
Given a depth map, a point cloud is derived in \cite{ye2011iccv} and used as the input data for their model. Firstly, the point cloud is cleaned and transformed into a canonical coordinate frame, allowing its matching against a set of predefined exemplars. This matching provides an initial estimation that is further refined in subsequent steps. Finally, the body pose is extracted from the improved point cloud.
In \citep{Hernandez2012}, random forest-based depth map labelling is combined with a probabilistic graphical model where graph-cut is applied for inference. Temporal constraints are used on sequences to improve the accuracy. 
In this work, we show that neither a pixel-level classification stage nor temporal information is needed to obtain the body pose.

Given a depth map, the proposal of \citep{Jiu2014} segments body parts by classifying each pixel independently  with a logistic regression function that receives as input ConvNet-based features. The ConvNet is initiallized by using an energy function that imposes spatial relations between body parts. However, an explicit body pose is not computed with their method.

In \citep{Madadi2015}, it is presented a multi-part body segmentation approach that is able to label each point of the depth map into body parts. A Histogram of Oriented Gradient descriptor is used to represent the body shape that will be matched against pose clusters learnt by Expectation Maximization. Then, the test pose is warped against the nearest cluster prototype to perform the part labelling.
Our proposal uses the idea of keeping a dictionary of prototype poses as well. However, in our case, the pose descriptor is automatically learnt by a ConvNet, and we do not need to select just a single prototype pose, as we combine all of them to obtain the final pose.
In \citep{Shafaei16} multiple depth cameras to estimate body joints are used. Their approach comprises a point-wise classification of depth maps by using a ConvNet; view aggregation of the independently classified views by point cloud reconstruction; and, body pose estimation by using linear regression on statistics computed from the point cloud.
In our model, we do not need either an intermediate representation (\ie~point cloud) or multiple cameras to obtain the body joints. We can deliver a body pose from a single camera viewpoint and, if multiple cameras are available, we can refine the estimated pose by integrating the poses obtained by each camera viewpoint.
The model proposed in \cite{haque2016eccv} uses local patches from depth data to detect body parts, initially obtaining a local representation of the body. Then, a convolutional and recurrent network (LSTM) are combined to iteratively obtain a global body pose.
In contrast, our model does not need either local patches or any iterative estimation, obtaining the 3D body pose in just a single step.

A voxel-to-voxel network (coined V2V-PoseNet) is proposed in \cite{moon2018cvpr} for hand and body pose estimation. Given a point cloud, the 3D space is split into a grid of voxels. For each voxel, the network estimates the likelihood of each body joint. The key components of the network are 3D convolutional blocks, an encoder and a decoder. In our case, we directly work on the depth maps and our architecture only used 2D convolutions what makes that operation faster.

With regard to disparity-based approaches, \citep{LopezQuintero2016} propose a pictorial structure model able to estimate 2D poses on stereo image pairs extracted from 3D videos. This model is extended in \citep{LopezQuintero2017} by adding temporal constraints and, therefore, applied to short 3D video sequences. However, none of these methods is able to output a full 3D pose.

Apart from the previously reviewed works that use either depth or disparity maps as input. There is an increasing number of works that directly process RGB images.
The model proposed in \cite{tekin2015arxiv} works on RGB images and requires a volume of aligned image windows of people to estimate the 3D body pose of the central one, where a 3D HoG descriptor is used as input of a regression method (Kernel Ridge Regression or Kernel Dependency Estimation). In contrast to our proposal, this method is limited to image sequences.

The concept of temporal consistency is added to the 3D Pictorial Structure model in \cite{belagiannis2014weccv} to address the problem of
multiple human pose estimation from multiple camera viewpoints. Later, some co-authors of that work propose in \cite{belagiannis2016mva} the estimation of body poses in operating rooms by combining ConvNet-based part detectors with the 3D Pictorial Structure model.
Although it is shown that the framework is able to deal with simultaneous people in the same scene, by using a person tracker, it is not able to estimate the body pose with a single viewpoint, as we propose in our work.

The model proposed in \cite{rogez2017lcr} is a multi-task ConvNet that simultaneously detects people and estimate their 2D and 3D body pose from RGB images. For that purpose, each input image region is classified into a set of predefined keyposes. Such keyposes are subsequently refined through regression. A set of candidate poses, with their associated scores, is generated for each person, allowing a post-processing stage to output a final pose per person. 
In our work, we also use a set of keyposes but with a different objective (\ie~to combine them).

In the recent work of \cite{katircioglu2018ijcv}, firstly, by using an autoencoder, the 3D body poses are embedded into high-dimensional space. Then, a mapping from RGB images to the previous space is learnt. And, finally, a LSTM model is used to process sequences of images, obtaining the final 3D pose.
We show in our work that a simpler method is also effective to obtain state-of-the-art results.


In summary, our end-to-end model targets depth maps and, in contrast to previous approaches, does not need pixel-wise segmentation as an intermediate representation; temporal information is not needed; and, the use of multiple cameras is optional, but helpful to refine the estimation. 

\section{Proposed model} \label{sec:model}
We propose a ConvNet, named `Deep Depth Pose' (DDP), that receives as input a depth map representing a person and returns the 3D pose of the person. 
Our model assumes that a body pose  $\mathbf{P}$ can be approximated by a linear combination of $K$ prototype poses $\mathbf{C}_i$:
\begin{equation}\label{eq:poselcombTeaser}
    \mathbf{P} = w_1 \cdot \mathbf{C}_1 + w_2 \cdot \mathbf{C}_2 + w_3 \cdot \mathbf{C}_3 + \ldots + w_K \cdot \mathbf{C}_K   \textrm{ },
\end{equation}
where $w_i$ is the weight assigned to the prototype $\mathbf{C}_i$.
Therefore, given a set of prototype poses learnt (\eg~by using K-means clustering) on the 3D space of body poses, DDP will output the set of weights that better approximates the corresponding 3D pose of its input. This idea is summarized in Fig.~\ref{fig:teaser}.
In this work we experiment with two body models (\ie~skeletons), described in Sec.~\ref{sec:expers}, corresponding to the ones proposed in the datasets used for evaluating our approach. 

In the following subsections, we start by  describing the architecture (\ie~layers) of the proposed DDP (Sec.~\ref{subsec:archit}) and, then, the loss functions defined for training it (Sec.~\ref{subsec:DDPtrain}). We continue by specifying how the 3D pose is obtained at test time (\ref{subsec:DDPinfer}). Finally, we show how we approach the multicamera case (Sec.~\ref{subsec:multiview}).

\begin{table}[t]
\caption{\textbf{DDP architecture.} Acronyms: `P'=pooling size; `Dr'=dropout; `$K$'=number of pose clusters. }
\label{tab:cnnarch}
\footnotesize
\begin{center}
\setlength{\tabcolsep}{0.2em} %
\begin{tabular}{|l|c|c|c|c|c|c|c|c|c|}
\hline 
\textit{Input} & Conv01 & Conv02 & Conv03 & Conv04 & Conv05 &Full01 & Full02 & Full03\\ 
\hline \hline
$100 \times 100$ & $7 \times 7  \times 96$ & $5 \times 5 \times 192$ & $3 \times 3 \times 512$ & $2 \times 2 \times 1024$ & $2 \times 2 \times 2048$ & 1024 & 256 & $K$\\
 & P: $2\times 2$ & P: $2\times 2$ & P: $2\times 2$ &  & & Dr=0.2  & & \\
\hline 
\end{tabular} 
\end{center}
\end{table}

\subsection{Model description}\label{subsec:archit}

The proposed DDP model is a ConvNet defined by the layers summarized in Tab.~\ref{tab:cnnarch}, where the input is a single depth map of size $100\times 100$ pixels and each convolutional block (\textit{Conv}) contains a ReLU, and a pooling layer (when indicated). For regularization purposes, dropout is used after the first fully-connected (\textit{Full01}) layer.

Note that the ideas described in the following subsections do not heavily depend on the specific architecture used. And, therefore they can be directly used with deeper networks (\eg~VGG\textit{x}~\cite{simonyan2014very} or ResNet~\cite{he2016resnet}), as long as the number of unit of the output layer is $K$. However, we show experimentally in this paper (Sec~\ref{sec:expers}) that state-of-the-art results can be obtained with the previously proposed architecture without the need of using pretrained networks as starting point or a huge amount of data (what usually happens for deeper networks).

\subsection{Model training}\label{subsec:DDPtrain}
%
%
\begin{figure*}[b]
\centering
   \includegraphics[width=0.98 \textwidth]{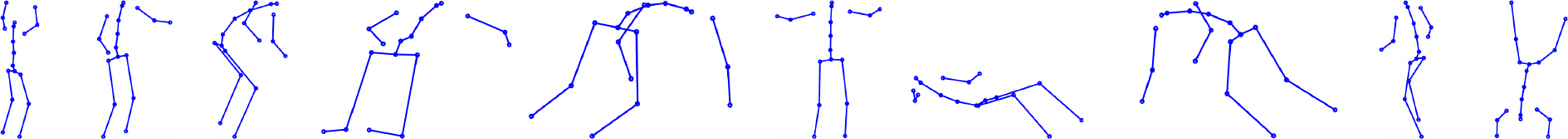}
   \caption{\textbf{Example of prototypes learnt during pose clustering on UBC3V dataset}. Note the variety of body limb configurations. Filled circles correspond to joints.
   }
   \label{fig:clusterPoses}
\end{figure*}
%
Let $\mathcal{C}$ be a set of $K$ cluster poses obtained by a clustering approach on the 3D space representation. And,
let $\mathcal{S}$ be a set of training pairs $\mathcal{S} = \left\{ (\mathbf{D}, \mathbf{P})  \right\}$,
where $\mathbf{D}$ and $\mathbf{P}$ are the depth map of the sample and its corresponding 3D pose, respectively. Where $\mathbf{P}$ is defined by $J$ joint positions. 
Our goal is to learn during training the DDP parameters ($\theta$), that minimize the loss function $\mathcal{L}_{\text{DDP}}$ defined as:

\begin{equation} \label{eq:mtLoss}
    \mathcal{L}_{\text{DDP}}\left( g(\mathbf{D}, \theta), \mathbf{C}, \mathbf{p}, \alpha \right) =
    (1-\alpha) \cdot \mathcal{L}_{\text{R}} \left( \mathbf{C} \times g(\mathbf{D}, \theta) , \mathbf{p} \right)
    + \alpha \cdot ||g(\mathbf{D}, \theta)||_1 
\end{equation}
This loss function comprises two parts: a residual loss $\mathcal{L}_{\text{R}}$ and a regularization term.
In the previous equation, $g(\mathbf{D}, \theta)$  represents a non-linear function on the input depth map $\mathbf{D}$ with parameters $\theta$ that returns a column vector of length $K$, $\mathbf{C}$ is a matrix with $K$ columns containing a prototype pose $c$ per column, $\mathbf{p}$ is the vectorized version of the ground-truth pose $\mathbf{P}$, $||\mathbf{z}||_1$ corresponds to the L1 norm of a target vector $\mathbf{z}$, and $\alpha$ is a  parameter to be selected during training to control the magnitude of the learnt weights of the prototype poses.

\myparagraph{Residual loss.}
Given a depth map containing a person, the role of the function $g(\mathbf{D}, \theta)$ is to estimate vector $\mathbf{w}$, \ie,~the weight of each pose cluster centroid $w_c$, subject to:
\begin{equation}\label{eq:poselcomb}
    \hat{\mathbf{p}} = \mathbf{C} \times \mathbf{w} = \sum_{i=1}^K w_i \cdot \mathbf{C}_{\cdot,i} \textrm{ },
\end{equation}
where $\hat{\mathbf{p}}$ represents the vectorized version of the estimated 3D pose of the target sample; $\mathbf{C}$ and $\mathbf{C}_{\cdot,i}$ are the matrix containing the set of prototype poses and the $i$-th column of matrix $\mathbf{C}$, respectively; and, $\mathbf{w}$ the column-vector containing in the $i$-th position the weight $w_i$  of the prototype $i$ in $\mathbf{C}$. 
In this work, $g(\mathbf{D}, \theta)$ is modelled as a ConvNet, where the last fully-connected layer contains $K$ units representing the weight of each prototype pose. Ideally, the estimated pose $\hat{\mathbf{p}}$ should be as similar as possible to the ground-truth pose $\mathbf{p}$.

Therefore, let $\mathbf{r}= \mathbf{p} - \hat{\mathbf{p}}$ be the residual vector obtained from the difference between the expected output pose $\mathbf{p}$ and the output pose $\hat{\mathbf{p}}$ obtained by applying the Eq.~\ref{eq:poselcomb}. 
Based on the L1-smooth norm, we can define the following loss function $\mathcal{L}_{\text{R}}$ :

\begin{equation}\label{eq:regLoss}
   \mathcal{L}_{\text{R}}(\mathbf{\hat{p}},\mathbf{p}) = \sum_{k=1}^{3J} ||r_k||_{S}
\end{equation}

\begin{equation}\label{eq:l1smoothnorm}
  ||r_{k}||_S=
      \left\{
                \begin{array}{lcl}
                  0.5 \cdot \sigma^2 \cdot r_{k}^2 &,& \mathrm{if} |r_{k}| < \frac{1}{\sigma^2} \\
                  |r_{k}| - \frac{0.5}{\sigma^2} &,& \mathrm{otherwise}
                \end{array}
              \right.
\end{equation}
where $r_{k}$ is the $k$-th component of vector $\mathbf{r}$, and $J$  the number of joints of the body model.

The L1-smooth norm is more robust to outliers than the classic L2 norm. Note that $\sigma^2$ is a parameter to be selected during training by cross-validation.

\subsection{Inference}\label{subsec:DDPinfer}
Given a test sample, firstly, it is resized to the DDP needed input size (see Sec.~\ref{subsec:impldet}). Then, the network will output (\ie~forward pass) the set of weights needed to obtain the pose of the input test sample by applying Eq.~\ref{eq:poselcomb}.

\subsection{Multiview estimation} \label{subsec:multiview}
Let us assume that we have $N$ camera viewpoints of the same scene. Therefore, we can compute $N$ point estimations $\mathbf{p}_j^v$ for the same joint $j$, where $v$ represents a camera index. If we have also information about the extrinsic parameters of the cameras, we can register all the estimated joints into a single reference system. A combined estimation $\mathbf{\bar{p}}_j$ of a body joint can be computed as:
\begin{equation}\label{eq:mcpt}
    \mathbf{\bar{p}}_j = \sum_v \omega_v \cdot \mathbf{p}_j^v \text{, s.t. } \sum_v \omega_v = 1
\end{equation}

A particular case of this equation is to assign the same weight to each camera, \ie, to average the poses. For example, in case we had three cameras, as we will see in the experimental section for `UBC3V' dataset, each camera will receive $\omega_v = 1/3$.

\section{Datasets} \label{sec:dataset}
We describe below the two datasets used to validate our model.

\subsection{ITOP} \label{subsec:dataitop}
\begin{figure*}[tb]
\centering
   \includegraphics[width=0.98 \textwidth]{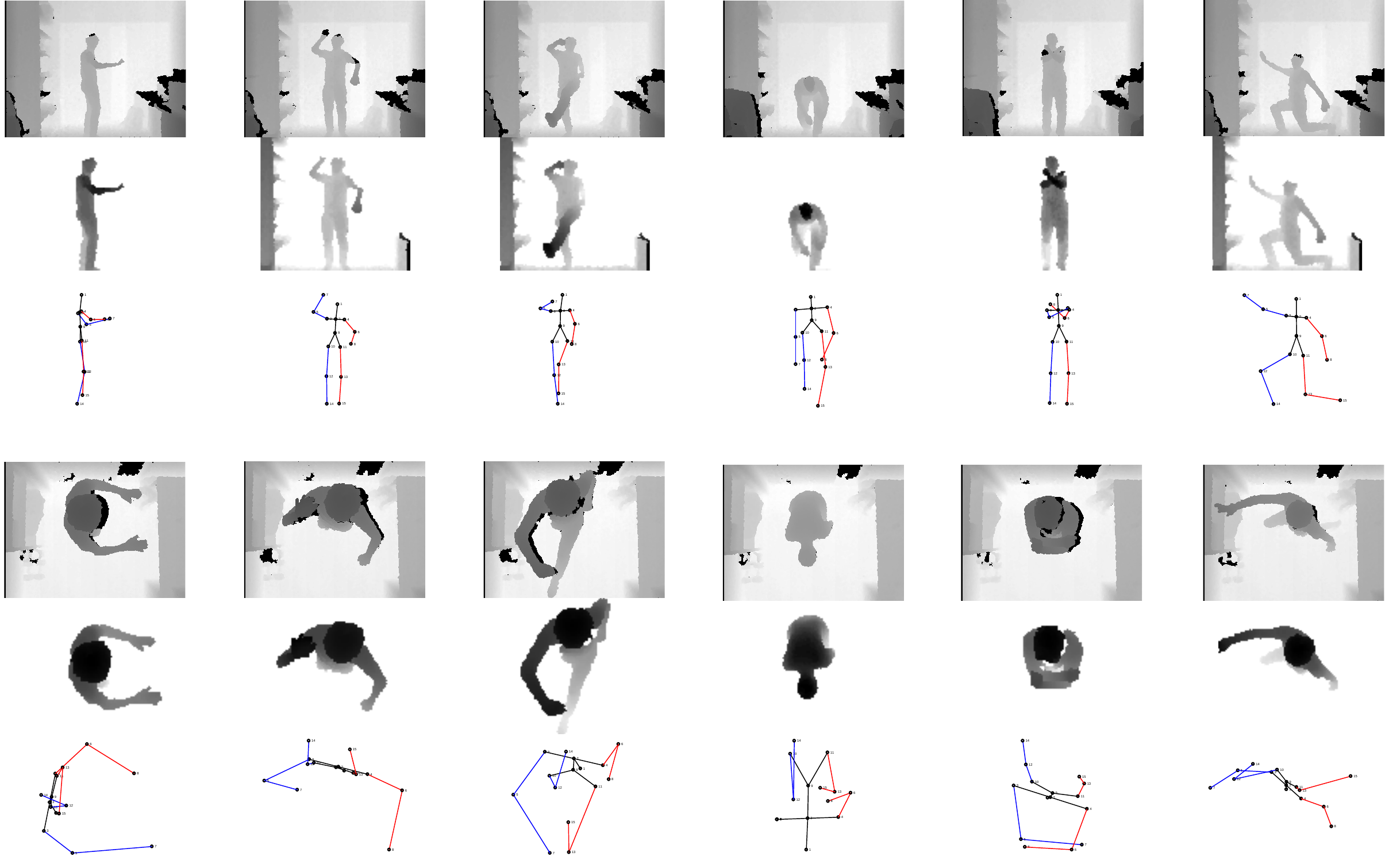}
   \caption{\textbf{Samples extracted from ITOP dataset}. Each block correspond to a camera viewpoint: \textit{frontal} and \textit{top}. Each block contains, from top to bottom: full-frame depth map ($320 \times 240$ pixels); pre-processed depth map (\ie~actual input for the network: $100\times 100$ pixels); and, ground-truth pose.}
   \label{fig:datasetITOP}
\end{figure*}

The Invariant-Top View Dataset (ITOP) \cite{haque2016eccv} contains 
depth images ($320\times 240$ pixels) collected by using two Asus Xtion PRO cameras, thus, providing two camera viewpoints. One of them corresponds to a top viewpoint, whereas the other to a frontal one. In ITOP 20 people carry out 15 actions each. 
The dataset is split into \textit{Training} (around $18k$ samples) and \textit{Test} samples (around 4800).
Some examples are shown in Fig.~\ref{fig:datasetITOP}, for both the frontal and top camera viewpoints. Note that the background is not plain and the depth maps are noisy, in contrast to the UBC3V dataset (see below). 
 The body model used in this dataset contains 15 joints, named: `Head', `Neck', `R-Shoulder', `L-Shoulder', `R-Elbow', `L-Elbow', `R-Hand', `L-Hand', `Torso', `R-Hip', `L-Hip', `R-Knee', `L-Knee', `R-Foot' and `L-Foot'.

\paragraph{Depth map pre-processing}
In the experiments where it is explicitly indicated, instead of using the original content of the depth maps, we apply a simple pre-processing pipeline to remove part of the background (as represented in Fig.~\ref{fig:datasetITOP}).
In particular, assuming that the person is approximately centred in the frame, we estimate the mean depth value of the person and use it to define an interval of depth values that allows applying a simple thresholding. Afterwards, morphological operations (\ie~\textit{dilate} followed by \textit{erode}) are used to smooth the segmented depth map. 
Alternatively, any region growing algorithm could be applied to filter out some background.
As shown in Fig.~\ref{fig:datasetITOP}, there are cases where the pre-processing stage removes some pixels belonging to the body, and cases where part of background objects are not removed. 

\subsection{UBC3V Hard-pose} \label{subsec:dataubc3v}
We carry out experiments on the recent dataset `UBC3V Hard-pose' \citep{Shafaei16}.
It contains synthetic samples of human poses simulating the intrinsic parameters of a Kinect-2 sensor. Three camera viewpoints are available for each sample, providing ground-truth information about the extrinsic parameters of the virtual cameras. Render resolution is set to $512 \times 424$ pixels. 
Depth maps along with their corresponding ground-truth 3D poses are provided for evaluation purposes. The body model used in this dataset contains 18 joints, as it can be seen in Fig.~\ref{fig:clusterPoses}.

\begin{figure*}[tb]
\centering
   \includegraphics[width=0.95 \textwidth]{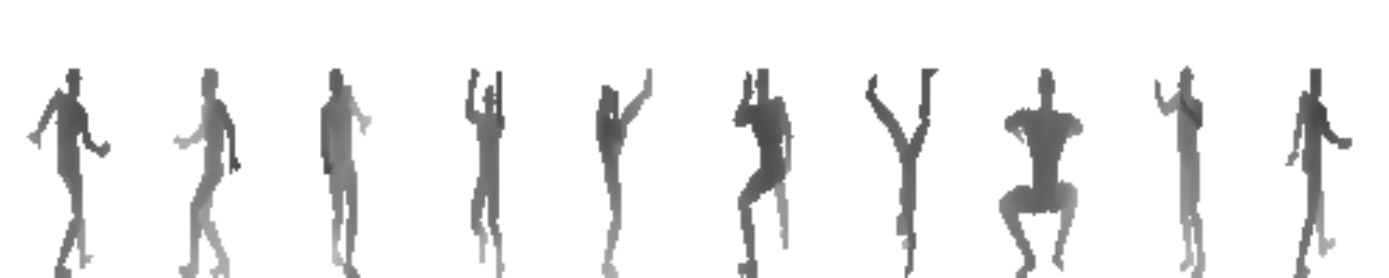}
   \caption{\textbf{Sample depth maps from UBC3V dataset}. From left to right, the first three samples correspond to the same pose but from three different camera viewpoints. Note the variety of challenging poses. }
   \label{fig:dataset}
\end{figure*}

As in \cite{Shafaei16}, we use the \textit{Test} set of the \textit{Hard-Pose} case to report our experimental results. This test set contains 19019 poses from three random camera viewpoints each, making a total of 57057 depth  frames. Models are previously trained on the 177177 samples of the \textit{Training} partition. In addition, 57057 extra samples are available for \textit{Validation}.
Some randomly selected samples from the Test Hard-Pose partition can be seen in Fig.~\ref{fig:dataset}. Note the variety of poses available in the dataset.

\section{Experiments} \label{sec:expers}
The objective of the following experiments is to validate the proposed model on the previously described datasets.

\subsection{Implementation details} \label{subsec:impldet}
The input to the DDP is a depth map resized to a size of $100 \times 100 $ pixels, as can be seen in Fig.~\ref{fig:datasetITOP}. 
For learning purposes, depth values are scaled to range $[0,1]$, and the joint coordinates to be learnt are normalized (\ie~zero mean and one standard deviation).

As the network is trained from the scratch, \ie, no pretrained model is used, the weights are initialized with random values following a normal distribution with mean zero and a standard deviation given by the `Xavier criterion' \cite{glorot2010xavier} (\ie~proportional to the filter size). Biases are initially set to zero.
Learning rate starts at $10^{-3}$ and is progressively reduced following a predefined protocol, with a maximum number of 1000 epochs. 

Our model is implemented in Matlab by using MatConvNet library~\citep{vedaldi2015matconvnet}. We plan to release our code and the pretrained models~\footnote{
Supplemental material: \url{http://www.uco.es/~in1majim/research/ddp.html}
}%
upon publication of this work. 

\subsection{Evaluation metrics} \label{subsec:metrics}
We use the following metrics to report experimental results:
\begin{itemize}
\setlength\itemsep{-0.1cm}
    \item Average error: it is defined as the average Euclidean distance between the estimated and ground-truth 3D locations of the joints. 
    \item Mean Average Precision (mAP): we could say that a target joint has been correctly estimated if its distance to the ground-truth is lower than a given threshold. Therefore, mAP gives information about the ratio of joints that have been correctly estimated given a threshold. The lower the threshold, the stricter this metric is. We will use a mAP curve to better understand the behaviour of the system.
    \item Area Under the Curve (AUC): by representing \textit{mAP} versus \textit{threshold} we obtain a curve that summarizes the behaviour of the model. The area under this curve is a compact representation of this behaviour. In an ideal situation, we would obtain a maximum value of 1. Therefore, the greater, the better. 
\end{itemize}

\subsection{Experimental results} \label{subsec:compstudy}
We define a set of experiments to evaluate the performance of our model.
%

\subsubsection{Baseline model}
%
\begin{figure}[t]
\centering
  \includegraphics[width=0.6 \textwidth]{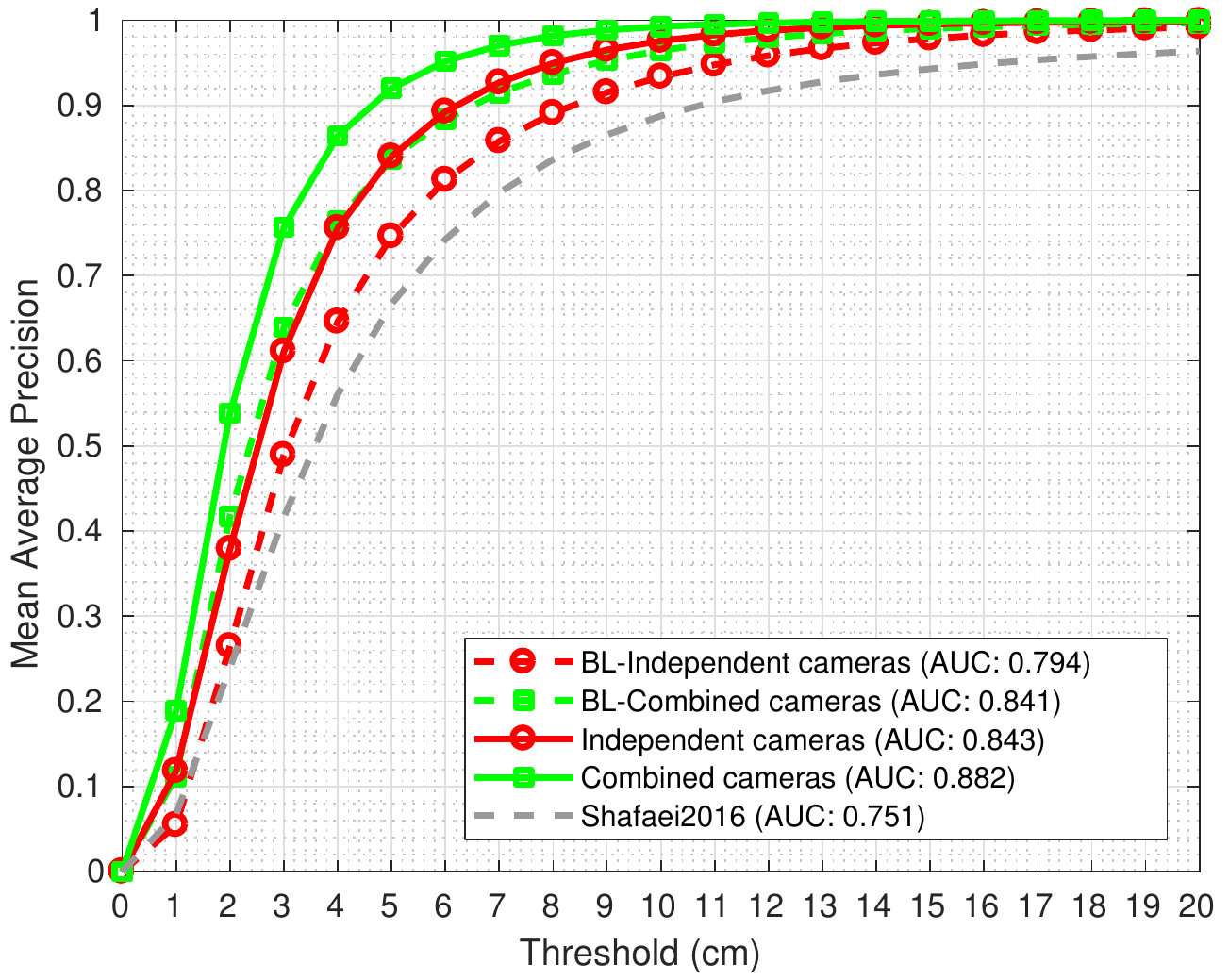}
  \caption{\textbf{UBC3V dataset: mean average precision at threshold.} Baseline (BL) is compared against our model, for single and multiview setups. Each curve represents the precision at a given threshold. The area-under-the-curve (AUC) for each case is included in the legend. The higher the better. Best viewed in color.}
  \label{fig:singleVSmulti}
\end{figure}
%
We define as \textit{baseline model} the direct regression of the body joint coordinates from the input depth map, without using a set of prototypes. For this purpose, we keep the very same configuration for all the layers of the DDP architecture proposed before but the last one, that is replaced by a fully-connected one with $3J$ units, depending on the number of joints $J$ of the target skeleton.

The results obtained by this baseline model (BL) on UBC3V are represented by dashed lines in Fig.~\ref{fig:singleVSmulti}. The dashed red line corresponds to considering each camera viewpoint as an independent sample. Whereas the dashed green line represents the results obtained by combining the three available camera viewpoints for each sample, as described in Sec.~\ref{subsec:multiview}.
The precision at 10 cm is $93.4$\% and $96.4$\% for the single and multiview cases, respectively.
The average error obtained for the single view case is $4.18$ cm, and $3.21$ cm for the multiview case. 
We will use these values for subsequent comparisons.

\begin{figure}[t]
\centering
  \includegraphics[width=0.6 \textwidth]{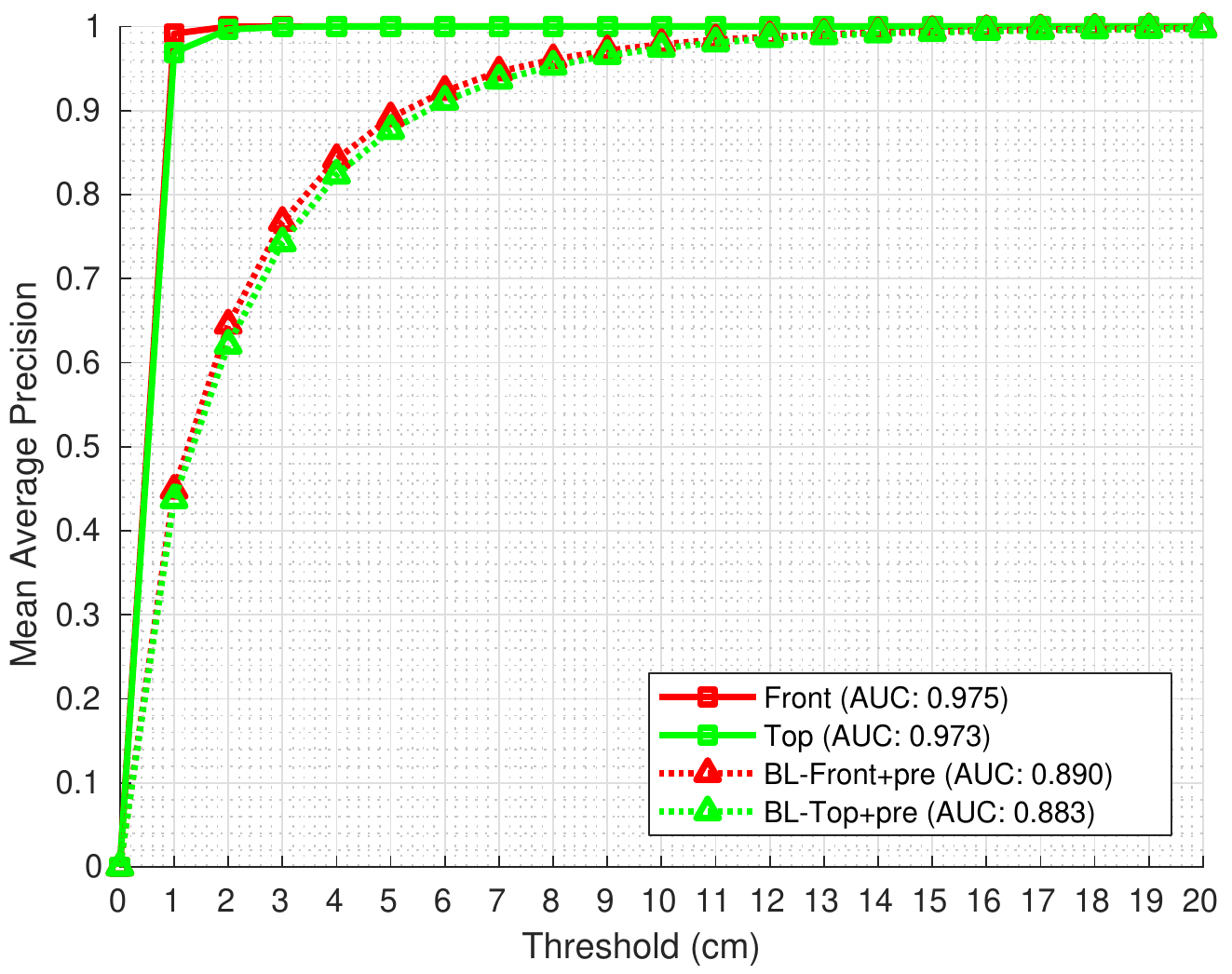}
  \caption{\textbf{ITOP dataset: mean average precision at threshold.} Each curve represents the precision at a given threshold. The area-under-the-curve (AUC) for each case is included in the legend. The higher the better. Best viewed in color.}
  \label{fig:curveITOP}
\end{figure}

The corresponding precision curves of the baseline model for ITOP are represented in Fig.~\ref{fig:curveITOP} (dashed lines). The input data has been pre-processed as previously described (\ie~foreground segmentation, Sec.~\ref{subsec:dataitop}). 
The precision at 10 cm is $97.9$\% and $97.4$\% for the frontal and top view cases, respectively.
The average error obtained for the frontal view case is $2.16$ cm, and $2.32$ cm for the top view one. 
%

\subsubsection{Ablation study of hyperparameters}
We explore here different values of the hyperparameters of the system: $K$ (\ie~the number of prototypes, Eq.~\ref{eq:poselcombTeaser}), $\sigma^2$ (\ie~threshold on the L1-smooth loss, Eq.~\ref{eq:l1smoothnorm}) and $\alpha$ (\ie~weight of the regularization term, Eq.~\ref{eq:mtLoss}).
With the help of the `Mann-Whitney U-test' ~\cite{MannWhitney1947test} we will select the values used for these hyperparameters.
By using the `Mann-Whitney U-test' we check if the median difference between pairs of configurations can be considered statistically significant (\ie~not by chance). We will report the $p$-value obtained for each pair of evaluated cases.
Note that, in order to ease the readability of the paper, all tables containing the statistical results of this set of experiments are placed at the end of the document as an Appendix.

\myparagraph{Number of prototypes.}
We fix 5 random seeds to initialize and train during 50 epochs models with different values of $K$, obtaining 5 trained models per number of prototypes.
In particular we explore the range $\left[10,120\right]$ in steps of 10.
Once training has finished, we compute the average loss on the validation set. 
In this experiment, the value of $\sigma^2$ has been fixed to $\sigma^2=1$, with $\alpha=0.01$, in all cases.

The results for UBC3V are summarized in Tab.~\ref{tab:statsKUBC3V}. If the $p$-value is not lower than $0.05$ we cannot assert that the medians of the loss values are different. For example, comparing the $K$ with the lowest loss, $K=100$, with the cases marked in bold we can only say that any of those cases would perform similarly as $K=100$. In our case, as any of them could be chosen expecting similar results, we will choose $K=100$ for the subsequent experiments on UBC3V.

The results for ITOP are summarized in Tables \ref{tab:statsKiTOPfview} and \ref{tab:statsKiTOPtview}. By analyzing both tables, and taking into account both the average loss and the $p$-values, we decide to choose $K=70$ for the subsequent experiments on ITOP.

\myparagraph{Value of $\sigma^2$ in L1-smooth loss.}
Following the same reasoning for the selection of $K$, in this case we select the value of the hyperparameter $\sigma^2$ of Eq.~\ref{eq:l1smoothnorm}. 
In this case, instead of using the validation loss as the guiding metric, we carry out our selection based on the mean squared error (MSE) of the vectorized pose coordinates (see Sec.~\ref{subsec:DDPtrain}) on the validation samples. The main reason for this choice is that the loss value changes with the value of $\sigma^2$ (\ie~the smaller the value of $\sigma^2$, the smaller the loss value).
The $p$-values obtained for UBC3V dataset are presented in Tab.~\ref{tab:statsS2UBC3V}, along with the average MSE per $\sigma^2$. We can see that statistically speaking, any value between $0.6$ and $2$ would be a good choice for $\sigma^2$. As $\sigma^2=0.8$ offers the lowest average error, we select that value for the remaining experiments on UBC3V.

Regarding ITOP dataset, based on the results summarized in Tables \ref{tab:statsS2iTOPfview} and \ref{tab:statsS2iTOPtview}, our choice is $\sigma^2=1$, for both camera viewpoints. Note that this value corresponds to the minimum average MSE for the frontal view. As there is no statistical difference with $\sigma^2=0.8$ in both camera viewpoints, choosing $\sigma^2=1$ is more convenient than other value as it helps to simplify Eq.~\ref{eq:l1smoothnorm}, reducing computational burden during training.

\myparagraph{Value of $\alpha$ in DDP loss.}
After fixing the values of $K$ and $\sigma^2$ for each dataset, we study here the effect of the hyperparameter $\alpha$, selecting its value for the final model.
The $p$-values obtained for UBC3V dataset are presented in Tab.~\ref{tab:statsAL1UBC3V}, along with the average MSE per $\alpha^2$. We can see that, the minimum MSE is obtained for $\alpha=0.01$. Note that any other of the explored values could have been chosen, as there is no statistical difference.

Regarding ITOP dataset, the results summarized in Tables~\ref{tab:statsAL1iTOPfview} and \ref{tab:statsAL1iTOPtview} indicate that a possible good choice for the weight of the regularization term is $\alpha = 0.08$. Note that, as in the previous cases, other values could be selected as well, expecting similar results.

\subsubsection{Single view pose estimation with DDP}
%
\begin{figure}[tb]
\centering
   \includegraphics[width=0.6 \textwidth]{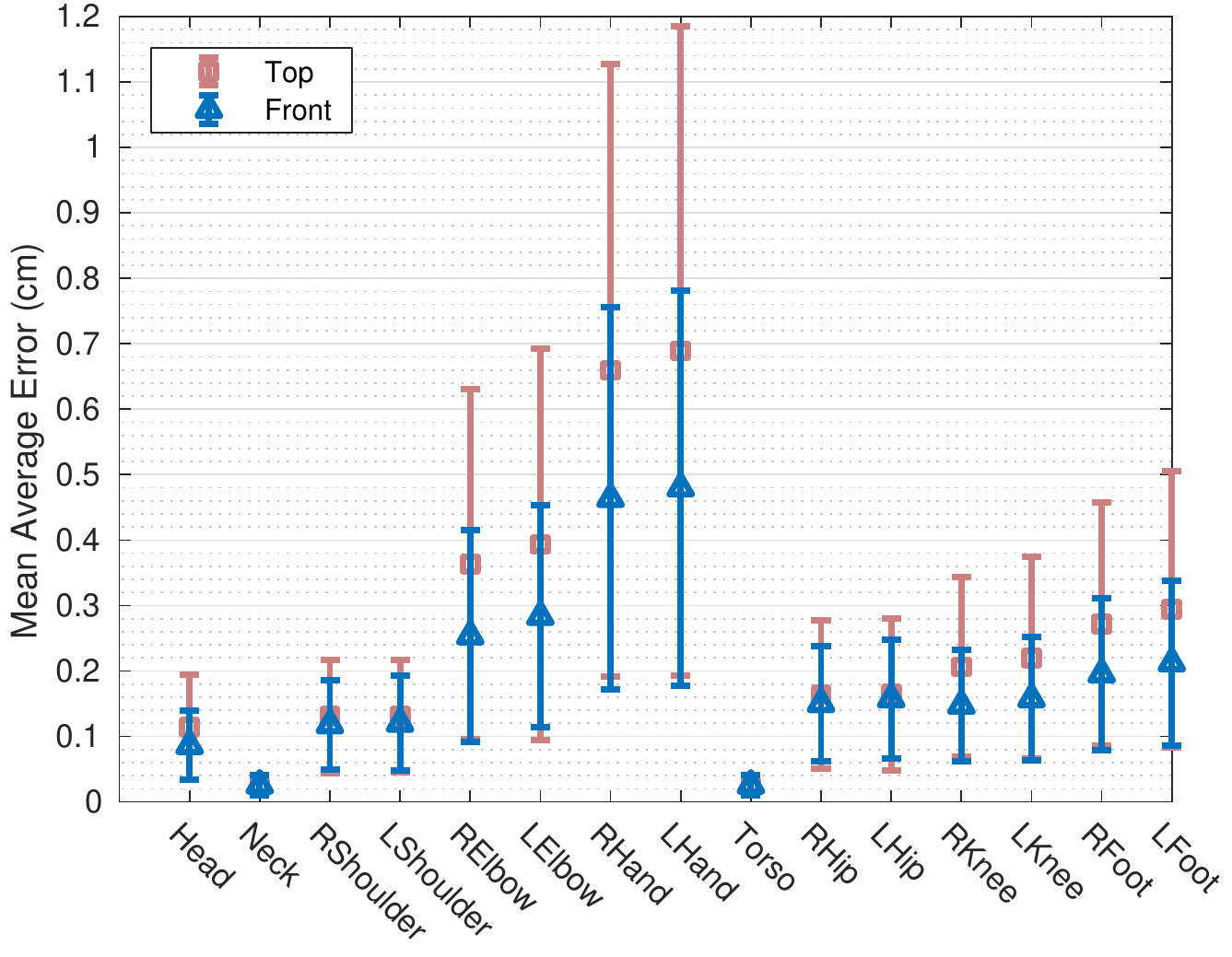}
   \caption{\textbf{ITOP: mean error per joint in cm (bars are one standard deviation).} Frontal and top camera viewpoints are compared. Best viewed in color.}
   \label{fig:jointsErrorCNNitop}
\end{figure}

\begin{figure}[tb]
\centering
   \includegraphics[width=0.6 \textwidth]{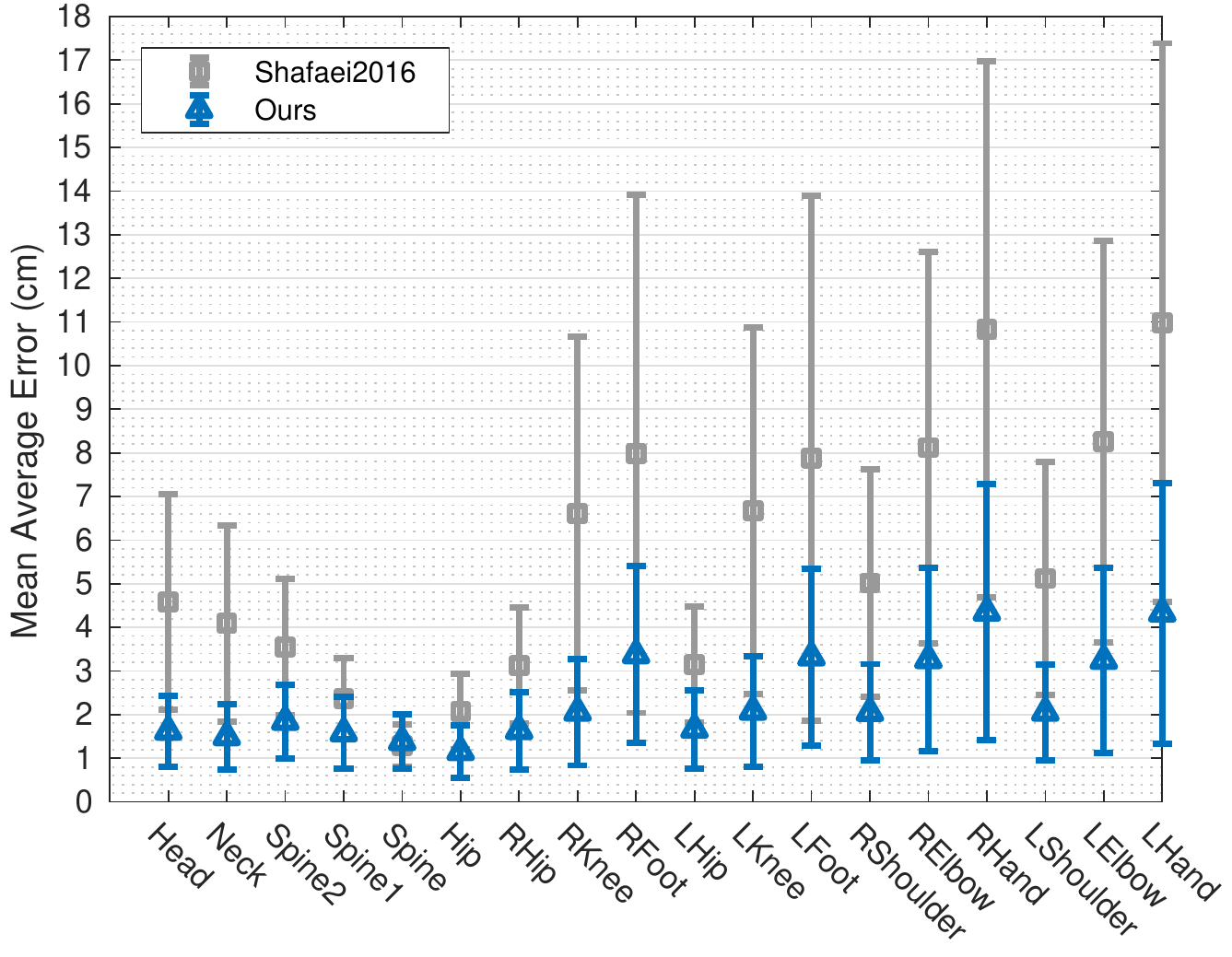}
   \caption{\textbf{UBC3V: mean error per joint in cm (bars are one standard deviation).} Our model compares favourably against Shafaei2016. In both cases, most errors are found in hands and feet. Best viewed in color.}
   \label{fig:jointsErrorCNN}
\end{figure}
%
For the UBC3V dataset, the solid red line of Fig.~\ref{fig:singleVSmulti} represents the mean average precision obtained for different thresholds, assuming that each camera viewpoint is an independent sample. Note that the AUC for this case clearly improves on the baseline model, even achieving a small improvement on the multiview case of the baseline (dashed green line). In terms of average error per limb the error is reduced from $4.18$ cm to $3.15$ cm (see Tab.~\ref{tab:resCompUBC3V}). This indicates that the proposed model is effective.

\begin{table}[tbh]
\caption{\textbf{Results on UBC3V.} Average error (cm), Precision at 10cm, and AUC are reported for different versions of the model. Acronyms: \textit{MC}=multicamera, \textit{Single}=independent viewpoints, \textit{BL}=baseline (no prototypes). 
}
\label{tab:resCompUBC3V}
\footnotesize
\begin{center}
\setlength{\tabcolsep}{0.2em} %
\begin{tabular}{|l|c|c|c|c|}
\hline 
\textit{Model} & Error & Prec@10 & AUC \\ 
\hline \hline
\textit{Ours-Single} &3.15 & 97.6 & 0.843\\
\textit{Ours-MC} &  \textbf{2.36} & \textbf{99.3} & \textbf{0.882}\\
\hline
\textit{BL-Single} & 4.18 & 93.4  & 0.794\\ 
\textit{BL-MC}  & 3.21 & 96.4 & 0.841\\ 
\hline
\textit{Shafaei16} & 5.64 & 88.7 & - \\
\hline 
\end{tabular} 
\end{center}
\end{table}

Focusing on the ITOP dataset, we can see in Fig.~\ref{fig:curveITOP} that the proposed model outperforms the baseline, in spite of using directly the raw depth data (\ie~non segmented). As the curves for the DDP model with the pre-processed inputs are quite similar to the ones without pre-processing, they have been omitted for clarity in this figure.
The actual values of average error, precision at 10cm and AUC are summarized in Tab.~\ref{tab:resCompITOP}. We report results for the full-body (FB), the upper-body (UB) and the lower-body (LB) joints separately.
See for example that the average errors of the baseline model are slightly greater than 2 cm, in contrast to the DDP ones that are around 10 times lower.
We can also see that the contribution of the pre-segmentation step is very small. Therefore, it could be omitted if the task does not require so much accuracy.
A breakdown of the error per limb is visually summarized in Fig.~\ref{fig:jointsErrorCNNitop}. On the one hand, we can see that the highest error is usually found in the arms. And, on the other hand, the error derived from the top view is generally higher than the frontal one.

\begin{table}[tbh]
\caption{\textbf{Results on ITOP.} Average error (cm), Precision at 10cm, and AUC are reported for different versions of the model. Acronyms: \textit{FB}=full-body; \textit{UB}=upper-body; \textit{LB}=lower-body; \textit{BL}=baseline (no prototypes); \textit{Pre}=presegmentation step; \textit{MC}=multi-camera. As in \cite{haque2016eccv}, upper-body consists of the head, neck, shoulders, elbows and hands. Best results at full-body level are marked in bold.
}
\label{tab:resCompITOP}
\scriptsize
\begin{center}
\setlength{\tabcolsep}{0.2em} %
\begin{tabular}{|l|c c c|c c c|c c c|c c c|}
\cline{2-13}
 \multicolumn{1}{c|}{}& \multicolumn{3}{c|}{\textit{Front-view}} & \multicolumn{3}{c|}{\textit{Top-view}} & \multicolumn{3}{c|}{\textit{Mixed-Single}} & \multicolumn{3}{c|}{\textit{Mixed-MC}} \\
\cline{2-13} 
\multicolumn{1}{c|}{} & Error & Prec@10 & AUC & Error & Prec@10 & AUC & Error & Prec@10 & AUC & Error & Prec@10 & AUC \\ 
\cline{2-13} \hline
\textit{Ours}-FB & \textbf{0.19} & 100 & \textbf{0.975} & 0.26 & 100 & 0.973 & 1.13 & 99.8 & 0.941 & 0.89 & 99.9 & 0.949\\
\multicolumn{1}{|c|}{UB} & 0.23 & 100 & - & 0.31 & 100 & - & 0.68 & 100 & - & 0.54 & 100 & -\\
\multicolumn{1}{|c|}{LB} & 0.15 & 100 & - & 0.19 & 100 & - & 1.55 & 99.7 & - & 1.29 & 99.9 & -\\
\hline
\textit{Ours}+Pre-FB & 0.25 & 100& 0.974& \textbf{0.23} & 100 & \textbf{0.974}&
\textbf{1.09} & 99.8 & \textbf{0.942} & \textbf{0.86} & 99.9 & \textbf{0.950}\\
\multicolumn{1}{|c|}{UB} & 0.29& 100&- & 0.28 & 100 & -&
0.61 &100&- & 0.49 & 100 & -\\
\multicolumn{1}{|c|}{LB} & 0.18& 100&- & 0.17 & 100 & - &
1.53 & 99.7 & - & 1.30 & 99.8 &-\\
\hline
\cline{1-7}
\textit{BL}+Pre-FB & 2.16 & 97.9 & 0.890 & 2.32 & 97.4 & 0.883 \\ 
\cline{1-7} 
\textit{Haque16}-FB & - & 77.4 & - & -  & 75.5 & -\\
\multicolumn{1}{|c|}{UB} & - & 84.0 & - & -  & 91.4 & -\\
\multicolumn{1}{|c|}{LB} & - & 67.3 & - & -  & 54.7 & -\\
\cline{1-7}
\textit{Moon18}-FB & - & 88.7 & - & -  & 83.4 & -\\
\cline{1-7}

\end{tabular} 
\end{center}
\end{table}

If instead of having a separated model for each viewpoint camera, what makes sense for many controlled scenarios, we have a single model where the camera viewpoint is unknown, we obtain the results summarized in the block of columns named `\textit{Mixed-Single}'. Note that in this case, the average error is slightly greater than 1 cm, what is still valid for many real applications. For the case of the mixed cameras, the set of cluster prototypes is simply obtained as the union of the frontal and top prototype sets, therefore, $K=140$.

Focusing on the two groups of joints (FB vs LB), we realize that, in general, the LB error is lower than the FB one for the view specific models. However, this observation changes for the mixed case, where the LB error is more than twice the FB one. This makes sense as the UB parts are twice visible in the depth maps than the LB ones (\ie~upper-body is visible both in frontal and top views), and this mixed model has to deal with all cases with just one set of parameters.

\begin{figure*}[tbhp]
\centering
   \includegraphics[width=0.98\textwidth]{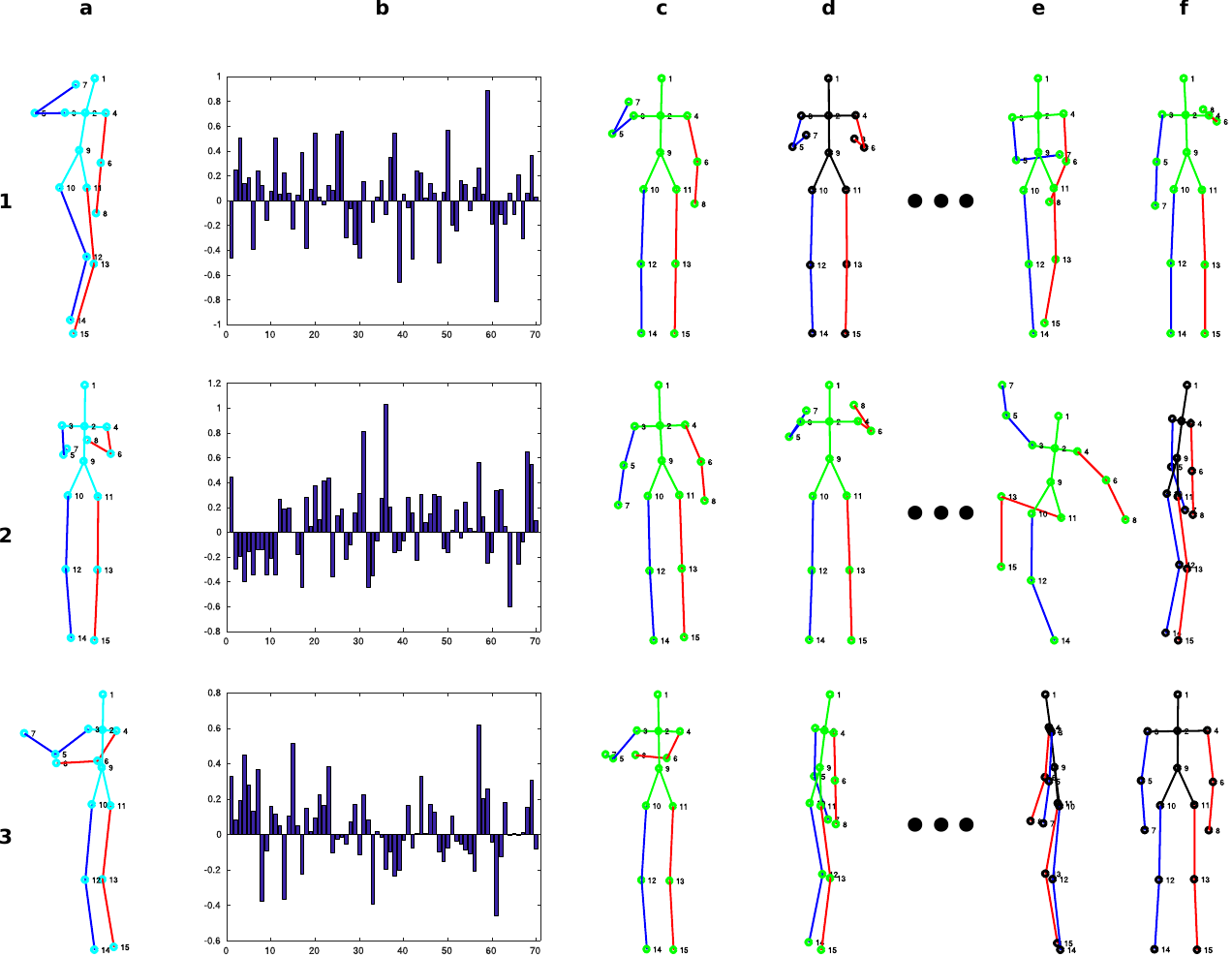}
   \caption{\textbf{Weights and prototype poses from ITOP}.
   \textbf{(a)} Target pose. \textbf{(b)} Weight estimated for each prototype pose ($K=70$). \textbf{(c)} to \textbf{(f)} Prototype poses sorted by decreasing absolute value of its weight.
   If the weight is positive, the trunk segments and the joints are drawn in green, otherwise, they are drawn in black. Best viewed in color.
   }
   \label{fig:weightsITOP}
\end{figure*}

In order to understand better the proposed model, we show in Fig.~\ref{fig:weightsITOP} the set of weights (column `b') generated by the DDP model to obtain the pose on the left (column `a'). For reference, we show both the two keyposes with the highest weights and with the lowest ones, in absolute value. For rows 1 and 3, the first prototype, in both cases (column `c'), corresponds to a pose visually similar to the target one. For the second row, we can easily see that the combination of the two first prototypes should approximate quite well the target pose. In contrast, the prototypes with the lowest weights, see for example row 2, are very different from the target pose.

\subsubsection{Contribution of multicamera estimation}
The goal of this experiment is to evaluate the contribution of the multi-camera fusion stage (see Sec.~\ref{subsec:multiview}). If it is not explicitly stated, each camera has been assigned the same weight $\omega_v$ for fusion.

The solid green line of Fig.~\ref{fig:singleVSmulti} represents the mean average precision obtained for different thresholds in UBC3V. 
The precision at 10cm is 99.3\% and its corresponding average error is 2.36cm. 
The AUC of this case is 0.882.
Note that these results improve both the BL and the single view case. In particular, AUC increases from $0.843$ to $0.882$. This suggests that combining cameras help to disambiguate hard situations generated by possible occluded body parts due to the viewpoint.
The mean error obtained for each body joint by the multiview system is summarized in Fig.~\ref{fig:jointsErrorCNN}. Major errors are localised in hands and feet, as they can be considered the most `deformable' parts of the model. While the lowest errors are localised in spine and hip. This behaviour is similar to the one reported in \citep{Shafaei16}.

For ITOP dataset, if we assume that for each test sample we have its corresponding frontal and top views, we can combine the corresponding pose estimations into a single one. Column \textit{`Mixed-MC'} of Tab.~\ref{tab:resCompITOP} summarizes the results obtained in such case. Comparing those results against the \textit{`Mixed-Single'}, if we focus on the \textit{Error} columns, we see that it is reduced from $1.13$ to $0.9$ and from $1.09$ to $0.86$, what is around $20\%$ error reduction in both cases. Note that single model is shared between the camera viewpoints ($K=140$).
For the case where pre-processing is not applied (first block of rows), it was found by cross-validation that assigning $\omega_{front}=0.6$ and $\omega_{top}=0.4$ obtains slightly better results than using $0.5$ for both cameras.
 
\subsubsection{Discussion of the results}
%
To put our results in context, we compare the results obtained by our model against the ones reported by the authors of the UBC3V dataset in \citep{Shafaei16}. The comparative results are summarized in Tab.~\ref{tab:resCompUBC3V}. 
Comparing our results (rows `Ours-\textit{x}') with \citep{Shafaei16}, we can observe that a new state-of-the-art has been established on this dataset.
In addition, it is reported in \citep{Shafaei16} that the error on \textit{groundtruth} is 2.44cm, and its precision at 10cm is 99.1\%, what indicates that our results are not very far from those \textit{groundtruth} values.

Regarding ITOP dataset, we have reached in both views $100\%$ precision at 10cm, what significantly surpasses the best published results~\cite{moon2018cvpr}, up to our knowledge, 88.7\% and 83.4\%.
It is worthwhile to mention that the baseline surprisingly achieves better results than the previously published results in both datasets, what might indicate that not always ``\textit{the more complex the model, the better}''.

\begin{figure*}[p]
\centering
   \includegraphics[width=0.85 \textwidth]{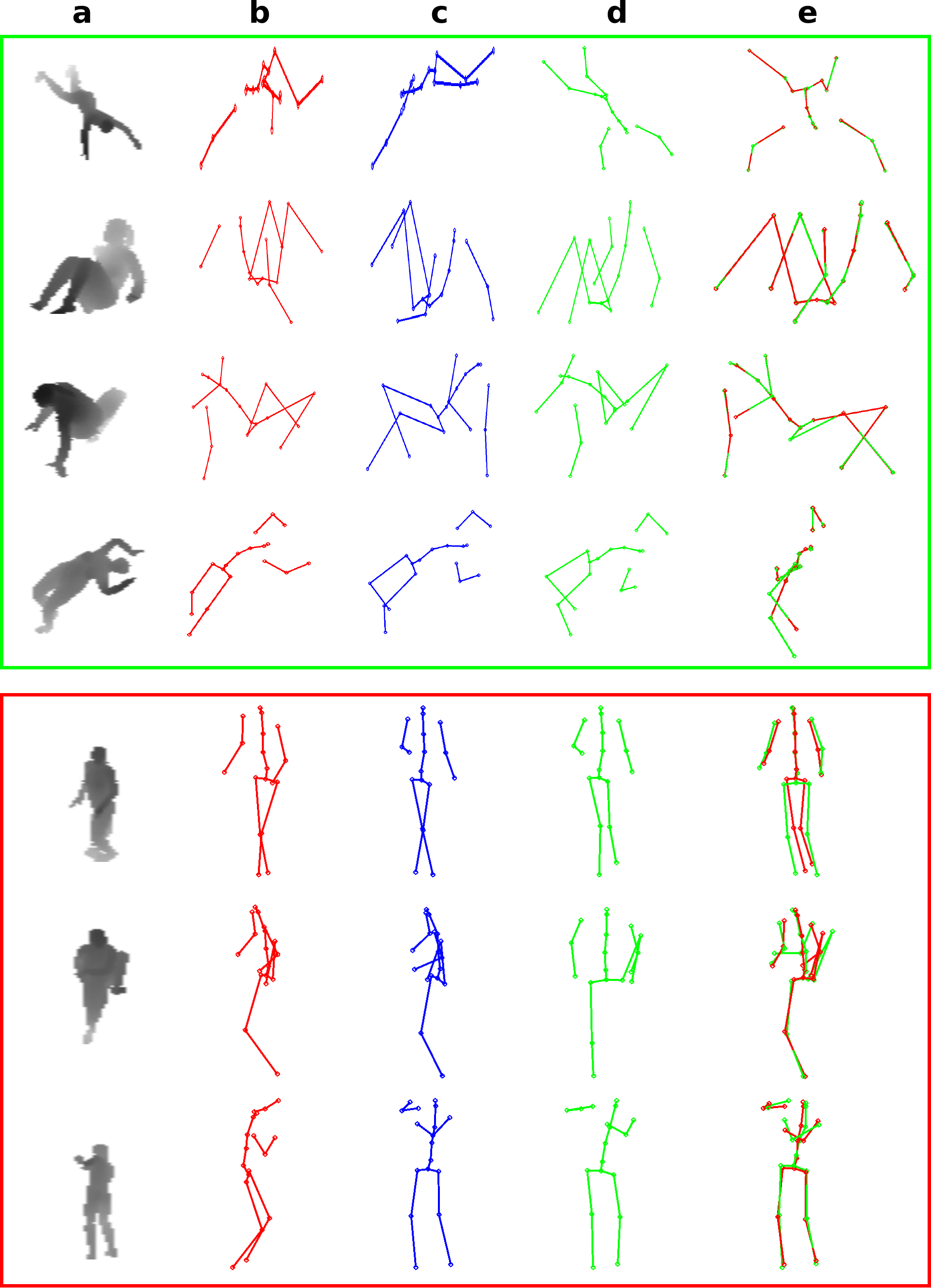}
   \caption{\textbf{Qualitative results on UBC3V}.
   \textbf{Top block}: successful cases (green rectangle). \textbf{Bottom block}: failure cases (red rectangle). Columns: (a) depth map from one camera; (b) Camera 1; (c) Camera 2; (d) Camera 3; (e) Multicamera pose (red) vs ground-truth (green). 
   Best viewed in color.
   }
   \label{fig:qresults}
\end{figure*}

\begin{figure*}[p]
\centering
   \includegraphics[width=0.98 \textwidth]{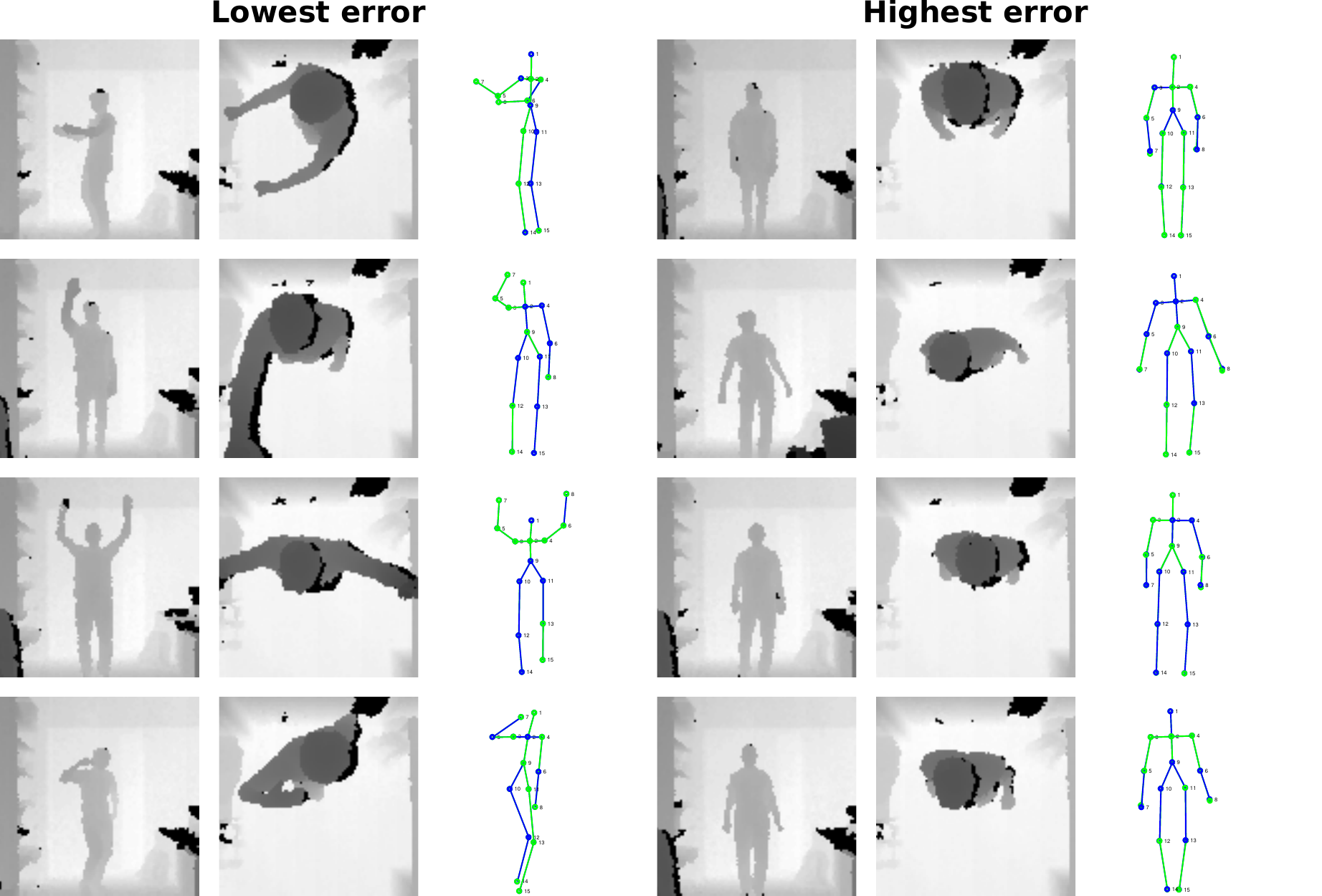}
   \caption{\textbf{Qualitative results on ITOP}.
   \textbf{Left block}: cases with low error. \textbf{Right block}: cases with higher error. Columns (from left to right): frontal  viewpoint depth map; top viewpoint depth map; and, estimated pose (blue) vs ground-truth (green). 
   Best viewed in digital format (zoom-in to find the  ``errorneous'' estimations).
   }
   \label{fig:qresultsITOP}
\end{figure*}

Regarding computational time, as a reference, in an  NVIDIA Titan Xp, for the ITOP models ($K=70$), the inference per sample takes around $0.16$ milliseconds, i.e., the full 3D pose of more than 6$k$ depth samples ($100\times100$ pixels) can be estimated in 1 second.

Some qualitative results are presented in Fig.~\ref{fig:qresults}, where column `a' contains an input depth maps from one camera, columns `b to `d' show the estimated poses per camera, and, column `e' compares the estimated multicamera pose (in red) against the ground-truth pose (in green).
Interestingly, we can observe that better results are obtained for unnatural poses than for the most common ones. Visually exploring the dataset, it seems that the proportion of `weird poses' is higher than the more natural ones. So, it makes sense that the model has focused on that type of poses. In addition, recall that the set of images that we are using is the so-called `Hard-Pose'. Therefore, depending on the target application it might be necessary to increase the number of underrepresented poses. Anyway, recall that the average error per limb is just $2.36$ cm, what can be valid for a wide range of applications.

With regard to ITOP dataset, from the results shown in Fig.~\ref{fig:qresultsITOP}, we can visually conclude that even the \textit{worst} estimations (right block) are fairly reasonable. Interestingly, it looks like that those cases are related to poses where the arms are in rest (\ie~close to the hips). What could mean that the system has put greater effort in learning the harder cases.

\section{Conclusions}\label{sec:conclus}
This paper has presented a new approach for 3D human pose estimation from depth maps. Our Deep Depth Model is able to deliver a 3D pose as a linear combination of prototype poses. It can be applied both to single camera viewpoints, and to multiple viewpoints, if the extrinsic camera parameters are available.
Limitations of previous approaches do not apply to our proposed model. Our method does not require either pixel-wise segmentation as an intermediate representation or temporal information to obtain the body pose, and the use of multiple cameras is only optional, but helpful to refine the estimated pose.

The experimental results on both ITOP and UBC3V datasets indicate that: 
(\textit{i}) using depth maps of resolution $100\times100$ pixels is enough to obtain a low average error ($<0.3$cm per joint on ITOP) in the pose estimation task, what can be good enough for many real applications;  
(\textit{ii}) the combination of multiple camera viewpoints helps to reduce the estimation error (\ie~from 3.15 cm to 2.36 cm on UBC3V); and (\textit{iii}), a new state-of-the-art has been established on both ITOP and UBC3V datasets, increasing the precision at 10cm around $16\%$ on ITOP and reducing the average error around $60\%$ (\ie~from 5.64 cm to 2.36 cm) on UBC3V.
%

\section*{Appendix} \label{sec:appendix}
This appendix contains the results obtained for the statistical tests carried out to select the hyper-parameters of the model.

Tables \ref{tab:statsKUBC3V}, \ref{tab:statsKiTOPfview} and \ref{tab:statsKiTOPtview} correspond to $K$ (\ie~number of prototypes) selection.

\begin{table}[t]
\caption{\textbf{$K$ selection for UBC3V dataset} $p$-values and average validation loss per $K$.}
\label{tab:statsKUBC3V}
\scriptsize
\begin{center}
\setlength{\tabcolsep}{0.2em} %
\begin{tabular}{|l|c c c c c c c c c c c c|}
\hline
 $K$ & \emph{10} & \emph{20}& \emph{30}& \emph{40}& \emph{50}& \emph{60}& \emph{70}& \emph{80}& \emph{90}& \emph{100}& \emph{110}& \emph{120} \\
 \hline \emph{10} & - & 0.0079 & 0.0079 & 0.0079 & 0.0079 & 0.0079 & 0.0079 & 0.0079 & 0.0079 & 0.0079 & 0.0079 & 0.0079  \\
   \emph{20} & 0.0079 & - & 0.0079 & 0.0079 & 0.0079 & 0.0079 & 0.0079 & 0.0079 & 0.0079 & 0.0079 & 0.0079 & 0.0079  \\
   \emph{30} & 0.0079 & 0.0079 & - & 0.0079 & 0.0079 & 0.0079 & 0.0079 & 0.0079 & 0.0079 & 0.0079 & 0.0079 & 0.0079  \\
   \emph{40} & 0.0079 & 0.0079 & 0.0079 & - & 0.3095 & 0.0079 & 0.0079 & 0.0079 & 0.0079 & 0.0079 & 0.0079 & 0.0079  \\
   \emph{50} & 0.0079 & 0.0079 & 0.0079 & 0.3095 & - & 0.0079 & 0.0079 & 0.0079 & 0.0079 & 0.0079 & 0.0079 & 0.0079  \\
   \emph{60} & 0.0079 & 0.0079 & 0.0079 & 0.0079 & 0.0079 & - & 0.6905 & 0.0556 & 0.1508 & 0.0556 & 0.0952 & 0.2222  \\
   \emph{70} & 0.0079 & 0.0079 & 0.0079 & 0.0079 & 0.0079 & 0.6905 & - & 0.0952 & 0.1508 & 0.0556 & 0.1508 & 0.3095  \\
   \emph{80} & 0.0079 & 0.0079 & 0.0079 & 0.0079 & 0.0079 & 0.0556 & 0.0952 & - & 0.6905 & 0.4206 & 1.0000 & 0.8413  \\
   \emph{90} & 0.0079 & 0.0079 & 0.0079 & 0.0079 & 0.0079 & 0.1508 & 0.1508 & 0.6905 & - & 0.5476 & 0.6905 & 1.0000  \\
   \emph{100} & 0.0079 & 0.0079 & 0.0079 & 0.0079 & 0.0079 & \textbf{0.0556} & \textbf{0.0556} & \textbf{0.4206} & \textbf{0.5476} & - & \textbf{0.1508} & \textbf{0.8413}  \\
   \emph{110} & 0.0079 & 0.0079 & 0.0079 & 0.0079 & 0.0079 & 0.0952 & 0.1508 & 1.0000 & 0.6905 & 0.1508 & - & 1.0000  \\
   \emph{120} & 0.0079 & 0.0079 & 0.0079 & 0.0079 & 0.0079 & 0.2222 & 0.3095 & 0.8413 & 1.0000 & 0.8413 & 1.0000 & -  \\
\hline 
\hline
\textit{Loss} & 5.2095   & 2.5667    &1.9072   & 1.5709    &1.5333    &1.3904    &1.3692   & 1.3219    &1.3077    &\textbf{1.3001}    &1.3336    &1.3346\\
\hline
\end{tabular} 
\end{center}
\end{table}

\begin{table}[t]
\caption{\textbf{$K$ selection for ITOP dataset (frontal view).} $p$-values and average validation loss per $K$.}
\label{tab:statsKiTOPfview}
\scriptsize
\begin{center}
\setlength{\tabcolsep}{0.2em} %
\begin{tabular}{|l|c c c c c c c c c c c c|}
\hline
 $K$ & \emph{10} & \emph{20}& \emph{30}& \emph{40}& \emph{50}& \emph{60}& \emph{70}& \emph{80}& \emph{90}& \emph{100}& \emph{110}& \emph{120} \\
 \hline
   \emph{10} & - & 0.0079 & 0.0079 & 0.0079 & 0.0079 & 0.0079 & 0.0079 & 0.0079 & 0.0079 & 0.0079 & 0.0079 & 0.0079  \\
   \emph{20} & 0.0079 & - & 0.0079 & 0.0079 & 0.0079 & 0.0079 & 0.0079 & 0.0079 & 0.0079 & 0.0079 & 0.0079 & 0.0079  \\
   \emph{30} & 0.0079 & 0.0079 & - & 0.0079 & 0.0079 & 0.0079 & 0.0079 & 0.0079 & 0.0079 & 0.0079 & 0.0079 & 0.0079  \\
   \emph{40} & 0.0079 & 0.0079 & 0.0079 & - & 0.0159 & 0.2222 & 0.0317 & 0.1508 & 0.0317 & 0.1508 & 0.6905 & 0.6905  \\
   \emph{50} & 0.0079 & 0.0079 & 0.0079 & 0.0159 & - & 0.1508 & 0.5476 & 0.1508 & 0.5476 & 0.1508 & 0.0079 & 0.0079  \\
   \emph{60} & 0.0079 & 0.0079 & 0.0079 & 0.2222 & 0.1508 & - & 0.0952 & 0.8413 & 0.0556 & 0.6905 & 0.0317 & 0.0317  \\
   \emph{70} & 0.0079 & 0.0079 & 0.0079 & 0.0317 & 0.5476 & 0.0952 & - & 0.0952 & 1.0000 & 0.1508 & 0.0159 & 0.0159  \\
   \emph{80} & 0.0079 & 0.0079 & 0.0079 & 0.1508 & 0.1508 & 0.8413 & 0.0952 & - & 0.0556 & 0.4206 & 0.0159 & 0.0556  \\
   \emph{90} & 0.0079 & 0.0079 & 0.0079 & 0.0317 & \textbf{0.5476} & \textbf{0.0556} & \textbf{1.0000} & \textbf{0.0556} & - & 0.0317 & 0.0079 & 0.0079  \\
   \emph{100} & 0.0079 & 0.0079 & 0.0079 & 0.1508 & 0.1508 & 0.6905 & 0.1508 & 0.4206 & 0.0317 & - & 0.0079 & 0.0079  \\
   \emph{110} & 0.0079 & 0.0079 & 0.0079 & 0.6905 & 0.0079 & 0.0317 & 0.0159 & 0.0159 & 0.0079 & 0.0079 & - & 1.0000  \\
   \emph{120} & 0.0079 & 0.0079 & 0.0079 & 0.6905 & 0.0079 & 0.0317 & 0.0159 & 0.0556 & 0.0079 & 0.0079 & 1.0000 & -  \\
\hline 
\hline
\textit{Loss} & 3.9688  &  1.6279  &0.9795 &   0.8172  &  0.7494&    0.7792&    0.7427&    0.7823 &   \textbf{0.7357} &    0.7700    &0.8313    &0.8474\\
\hline
\end{tabular} 
\end{center}
\end{table}

\begin{table}[t]
\caption{\textbf{$K$ selection for ITOP dataset (top view).} $p$-values and average validation loss per $K$.}
\label{tab:statsKiTOPtview}
\scriptsize
\begin{center}
\setlength{\tabcolsep}{0.2em} %
\begin{tabular}{|l|c c c c c c c c c c c c|}
\hline
 $K$ & \emph{10} & \emph{20}& \emph{30}& \emph{40}& \emph{50}& \emph{60}& \emph{70}& \emph{80}& \emph{90}& \emph{100}& \emph{110}& \emph{120} \\
 \hline
   \emph{10} & - & 0.0079 & 0.0079 & 0.0079 & 0.0079 & 0.0079 & 0.0079 & 0.0079 & 0.0079 & 0.0079 & 0.0079 & 0.0079  \\
   \emph{20} & 0.0079 & - & 0.0079 & 0.0079 & 0.0079 & 0.0079 & 0.0079 & 0.0079 & 0.0079 & 0.0079 & 0.0079 & 0.0079  \\
   \emph{30} & 0.0079 & 0.0079 & - & 0.0079 & 0.0079 & 0.0079 & 0.0079 & 0.0079 & 0.0159 & 0.2222 & 0.5476 & 0.6905  \\
   \emph{40} & 0.0079 & 0.0079 & 0.0079 & - & 0.0556 & 0.8413 & 0.2222 & 0.8413 & 0.0556 & 0.0952 & 0.0317 & 0.0317  \\
   \emph{50} & 0.0079 & 0.0079 & 0.0079 & 0.0556 & - & 0.0317 & 1.0000 & 0.3095 & 0.0079 & 0.0079 & 0.0079 & 0.0079  \\
   \emph{60} & 0.0079 & 0.0079 & 0.0079 & 0.8413 & 0.0317 & - & 0.3095 & 0.8413 & 0.0317 & 0.0952 & 0.0317 & 0.0317  \\
   \emph{70} & 0.0079 & 0.0079 & 0.0079 & \textbf{0.2222} & \textbf{1.0000} & \textbf{0.3095} & - & \textbf{0.4206} & \textbf{0.0952} & \textbf{0.0556} & 0.0159 & 0.0159  \\
   \emph{80} & 0.0079 & 0.0079 & 0.0079 & 0.8413 & 0.3095 & 0.8413 & 0.4206 & - & 0.0556 & 0.0952 & 0.0159 & 0.0159  \\
   \emph{90} & 0.0079 & 0.0079 & 0.0159 & 0.0556 & 0.0079 & 0.0317 & 0.0952 & 0.0556 & - & 1.0000 & 0.2222 & 0.1508  \\
   \emph{100} & 0.0079 & 0.0079 & 0.2222 & 0.0952 & 0.0079 & 0.0952 & 0.0556 & 0.0952 & 1.0000 & - & 0.3095 & 0.3095  \\
   \emph{110} & 0.0079 & 0.0079 & 0.5476 & 0.0317 & 0.0079 & 0.0317 & 0.0159 & 0.0159 & 0.2222 & 0.3095 & - & 0.8413  \\
  \emph{120} & 0.0079 & 0.0079 & 0.6905 & 0.0317 & 0.0079 & 0.0317 & 0.0159 & 0.0159 & 0.1508 & 0.3095 & 0.8413 & -  \\

\hline 
\hline
\textit{Loss} & 4.2101  &  1.7362 &   1.1705    &1.0432&    0.9791&    1.0398&    \textbf{0.9707}    &1.0235 &   1.1010  &  1.1104 &   1.1489  &  1.1627\\
\hline
\end{tabular} 
\end{center}
\end{table}

Tables \ref{tab:statsS2UBC3V}, \ref{tab:statsS2iTOPfview} and \ref{tab:statsS2iTOPtview} correspond to $\sigma^2$ (\ie~threshold for smooth-L1) selection.

\begin{table}[t]
\caption{\textbf{$\sigma^2$ selection for UBC3V dataset.} $p$-values and average validation MSE per $\sigma^2$.}
\label{tab:statsS2UBC3V}
\scriptsize 
\begin{center}
\setlength{\tabcolsep}{0.2em} %
\begin{tabular}{|l|c c c c c c c c |}
\hline
 $\sigma^2$ & \emph{0.2} & \emph{0.4}& \emph{0.6}& \emph{0.8}& \emph{1}& \emph{1.2}& \emph{1.5}& \emph{2} \\
  \hline \emph{0.2} & - & 0.8413 & 0.1508 & 0.0317 & 0.0556 & 0.3095 & 0.0317 & 0.0952  \\
  \hline \emph{0.4} & 0.8413 & - & 0.0556 & 0.0079 & 0.0317 & 0.1508 & 0.0079 & 0.0556  \\
  \hline \emph{0.6} & 0.1508 & 0.0556 & - & 0.5476 & 1.0000 & 1.0000 & 0.5476 & 0.5476  \\
  \hline \emph{0.8} & 0.0317 & 0.0079 & \textbf{0.5476} & - & \textbf{0.5476} & \textbf{0.6905} & \textbf{0.6905} & \textbf{0.3095}  \\
  \hline \emph{1} & 0.0556 & 0.0317 & 1.0000 & 0.5476 & - & 1.0000 & 1.0000 & 0.6905  \\
  \hline \emph{1.2} & 0.3095 & 0.1508 & 1.0000 & 0.6905 & 1.0000 & - & 1.0000 & 0.8413  \\
  \hline \emph{1.5} & 0.0317 & 0.0079 & 0.5476 & 0.6905 & 1.0000 & 1.0000 & - & 0.3095  \\
  \hline \emph{2} & 0.0952 & 0.0556 & 0.5476 & 0.3095 & 0.6905 & 0.8413 & 0.3095 & -  \\
\hline 
\hline
\textit{MSE} & 1.4035  &  1.4026  &  1.3359 &    \textbf{1.3092}&    1.3302&    1.3419&    1.3150&    1.3447\\
\hline
\end{tabular} 
\end{center}
\end{table}

\begin{table}[t]
\caption{\textbf{$\sigma^2$ selection for ITOP dataset (frontal view).} $p$-values and average validation MSE per $\sigma^2$.}
\label{tab:statsS2iTOPfview}
\scriptsize 
\begin{center}
\setlength{\tabcolsep}{0.2em} %
\begin{tabular}{|l|c c c c c c c c c c |}
\hline
 $\sigma^2$ & \emph{0.2} & \emph{0.4}& \emph{0.6}& \emph{0.8}& \emph{1}& \emph{1.2}& \emph{1.4}& \emph{1.6}& \emph{1.8}& \emph{2} \\
\hline \emph{0.2} & - & 1.0000 & 0.8413 & 0.8413 & \textbf{0.8413} & 0.6905 & 1.0000 & 0.8413 & 0.8413 & 1.0000  \\
  \hline \emph{0.4} & 1.0000 & - & 0.8413 & 0.3095 & \textbf{1.0000} & 0.8413 & 0.6905 & 0.3095 & 0.3095 & 0.8413  \\
  \hline \emph{0.6} & 0.8413 & 0.8413 & - & 0.6905 & \textbf{0.8413} & 0.4206 & 1.0000 & 0.2222 & 0.6905 & 1.0000  \\
  \hline \emph{0.8} & 0.8413 & 0.3095 & 0.6905 & - & \textbf{0.2222} & 1.0000 & 0.8413 & 0.8413 & 0.8413 & 0.6905  \\
  \hline \emph{1} & 0.8413 & 1.0000 & 0.8413 & 0.2222 & - & 0.5476 & 0.5476 & 0.5476 & 0.2222 & 0.8413  \\
  \hline \emph{1.2} & 0.6905 & 0.8413 & 0.4206 & 1.0000 & \textbf{0.5476} & - & 0.8413 & 0.6905 & 1.0000 & 0.8413  \\
  \hline \emph{1.4} & 1.0000 & 0.6905 & 1.0000 & 0.8413 & \textbf{0.5476} & 0.8413 & - & 0.6905 & 0.4206 & 0.8413  \\
  \hline \emph{1.6} & 0.8413 & 0.3095 & 0.2222 & 0.8413 & \textbf{0.5476} & 0.6905 & 0.6905 & - & 1.0000 & 0.5476  \\
  \hline \emph{1.8} & 0.8413 & 0.3095 & 0.6905 & 0.8413 & \textbf{0.2222} & 1.0000 & 0.4206 & 1.0000 & - & 0.8413  \\
  \hline \emph{2} & 1.0000 & 0.8413 & 1.0000 & 0.6905 & \textbf{0.8413} & 0.8413 & 0.8413 & 0.5476 & 0.8413 & -  \\

\hline 
\hline
\textit{MSE} & 0.7650  &  0.7573  &  0.7572&    0.7707&    \textbf{0.7517}&    0.7613&    0.7609&    0.7697  &  0.7719 &   0.7629\\
\hline
\end{tabular} 
\end{center}
\end{table}

\begin{table}[t]
\caption{\textbf{$\sigma^2$ selection for ITOP dataset (top view).} $p$-values and average validation MSE per $\sigma^2$.}
\label{tab:statsS2iTOPtview}
\scriptsize 
\begin{center}
\setlength{\tabcolsep}{0.2em} %
\begin{tabular}{|l|c c c c c c c c c c |}
\hline
 $\sigma^2$ & \emph{0.2} & \emph{0.4}& \emph{0.6}& \emph{0.8}& \emph{1}& \emph{1.2}& \emph{1.4}& \emph{1.6}& \emph{1.8}& \emph{2} \\
\hline \emph{0.2} & - & 0.8413 & 1.0000 & \textbf{0.1508} & 1.0000 & 0.8413 & 0.3095 & 0.4206 & 0.6905 & 0.2222  \\
  \hline \emph{0.4} & 0.8413 & - & 1.0000 & \textbf{0.0952} & 0.8413 & 0.8413 & 0.4206 & 0.4206 & 0.6905 & 0.2222  \\
  \hline \emph{0.6} & 1.0000 & 1.0000 & - & \textbf{0.4206} & 1.0000 & 1.0000 & 0.5476 & 0.6905 & 1.0000 & 0.3095  \\
  \hline \emph{0.8} & 0.1508 & 0.0952 & 0.4206 & - & 0.1508 & 0.3095 & 0.0556 & 0.0556 & 0.2222 & 0.0317  \\
  \hline \emph{1} & 1.0000 & 0.8413 & 1.0000 & \textbf{0.1508} & - & 1.0000 & 0.4206 & 0.5476 & 0.8413 & 0.2222  \\
  \hline \emph{1.2} & 0.8413 & 0.8413 & 1.0000 & \textbf{0.3095} & 1.0000 & - & 0.5476 & 0.6905 & 0.8413 & 0.3095  \\
  \hline \emph{1.4} & 0.3095 & 0.4206 & 0.5476 & \textbf{0.0556} & 0.4206 & 0.5476 & - & 1.0000 & 0.5476 & 0.8413  \\
  \hline \emph{1.6} & 0.4206 & 0.4206 & 0.6905 & \textbf{0.0556} & 0.5476 & 0.6905 & 1.0000 & - & 0.8413 & 0.5476  \\
  \hline \emph{1.8} & 0.6905 & 0.6905 & 1.0000 & \textbf{0.2222} & 0.8413 & 0.8413 & 0.5476 & 0.8413 & - & 0.3095  \\
  \hline \emph{2} & 0.2222 & 0.2222 & 0.3095 & 0.0317 & 0.2222 & 0.3095 & 0.8413 & 0.5476 & 0.3095 & -  \\
\hline 
\hline
\textit{MSE} & 1.0095  &  1.0166  &  1.0080&    \textbf{0.9738}&    1.0144&    1.0148 &   1.0394    &1.0344 &   1.0173 &   1.0478\\
\hline
\end{tabular} 
\end{center}
\end{table}

Tables \ref{tab:statsAL1UBC3V}, \ref{tab:statsAL1iTOPfview} and \ref{tab:statsAL1iTOPtview} correspond to $\alpha$ (\ie~regularization term weight) selection.

\begin{table}[t]
\caption{\textbf{$\alpha$ selection for UBC3V dataset.} $p$-values and average validation MSE per $\alpha$.}
\label{tab:statsAL1UBC3V}
\scriptsize 
\begin{center}
\setlength{\tabcolsep}{0.2em} %
\begin{tabular}{|l|c c c c c c c c c c|}
\hline
 $\alpha$ & \emph{0} &\emph{0.01} & \emph{0.02} & \emph{0.04}&\emph{0.05} & \emph{0.06}& \emph{0.08}& \emph{0.1}& \emph{0.2}& \emph{0.5}\\
\hline \emph{0} & - & \textbf{0.8413} & 0.8413 & 0.8413 & 0.8413 & 0.8413 & 1.0000 & 0.8413 & 0.6905 & 1.0000  \\
  \hline \emph{0.01} & 0.8413 & - & 0.0556 & 0.6905 & 1.0000 & 0.1508 & 1.0000 & 0.0952 & 0.4206 & 0.5476  \\
  \hline \emph{0.02} & 0.8413 & \textbf{0.0556} & - & 0.0317 & 0.1508 & 0.4206 & 0.1508 & 0.6905 & 0.5476 & 0.5476  \\
  \hline \emph{0.04} & 0.8413 & \textbf{0.6905} & 0.0317 & - & 0.2222 & 0.0952 & 0.3095 & 0.1508 & 0.6905 & 0.6905  \\
  \hline \emph{0.05} & 0.8413 & \textbf{1.0000} & 0.1508 & 0.2222 & - & 0.1508 & 0.8413 & 0.1508 & 0.4206 & 0.5476  \\
  \hline \emph{0.06} & 0.8413 & \textbf{0.1508} & 0.4206 & 0.0952 & 0.1508 & - & 0.1508 & 0.6905 & 1.0000 & 1.0000  \\
  \hline \emph{0.08} & 1.0000 & \textbf{1.0000} & 0.1508 & 0.3095 & 0.8413 & 0.1508 & - & 0.2222 & 0.5476 & 0.8413  \\
  \hline \emph{0.1} & 0.8413 & \textbf{0.0952} & 0.6905 & 0.1508 & 0.1508 & 0.6905 & 0.2222 & - & 0.5476 & 0.5476  \\
  \hline \emph{0.2} & 0.6905 & \textbf{0.4206} & 0.5476 & 0.6905 & 0.4206 & 1.0000 & 0.5476 & 0.5476 & - & 0.8413  \\
  \hline \emph{0.5} & 1.0000 & \textbf{0.5476} & 0.5476 & 0.6905 & 0.5476 & 1.0000 & 0.8413 & 0.5476 & 0.8413 & -  \\
\hline 
\hline
\textit{MSE} &  1.3346  &   \textbf{1.3092}  &  1.3462  &  1.3233 &   1.3182 &   1.3337 &   1.3267  &  1.3428 &    1.3441 &    1.3373\\
\hline
\end{tabular} 
\end{center}
\end{table}

\begin{table}[t]
\caption{\textbf{$\alpha$ selection for ITOP dataset (frontal view).} $p$-values and average validation MSE per $\alpha$.}
\label{tab:statsAL1iTOPfview}
\scriptsize 
\begin{center}
\setlength{\tabcolsep}{0.2em} %
\begin{tabular}{|l|c c c c c c c c |}
\hline
 $\alpha$ &\emph{0}& \emph{0.01} & \emph{0.02}& \emph{0.04}& \emph{0.05}& \emph{0.06}& \emph{0.08}& \emph{0.1} \\
  \hline \emph{0} & - & 1.0000 & 0.6905 & 0.8413 & 0.6905 & 0.8413 & \textbf{0.1508} & 0.8413  \\
  \hline \emph{0.01} & 1.0000 & - & 0.4206 & 0.6905 & 0.5476 & 0.6905 & \textbf{0.3095} & 1.0000  \\
  \hline \emph{0.02} & 0.6905 & 0.4206 & - & 0.3095 & 0.6905 & 0.6905 & \textbf{0.0556} & 0.5476  \\
  \hline \emph{0.04} & 0.8413 & 0.6905 & 0.3095 & - & 0.8413 & 1.0000 & \textbf{0.0952} & 1.0000  \\
  \hline \emph{0.05} & 0.6905 & 0.5476 & 0.6905 & 0.8413 & - & 1.0000 & \textbf{0.0952} & 0.6905  \\
  \hline \emph{0.06} & 0.8413 & 0.6905 & 0.6905 & 1.0000 & 1.0000 & - & \textbf{0.4206} & 0.6905  \\
  \hline \emph{0.08} & 0.1508 & 0.3095 & 0.0556 & 0.0952 & 0.0952 & 0.4206 & - & 0.6905  \\
  \hline \emph{0.1} & 0.8413 & 1.0000 & 0.5476 & 1.0000 & 0.6905 & 0.6905 & \textbf{0.6905} & -  \\
\hline 
\hline
\textit{MSE} & 0.7551 &   0.7517&    0.7641   & 0.7540 &   0.7579&    0.7563    &\textbf{0.7339} &   0.7510\\
\hline
\end{tabular} 
\end{center}
\end{table}

\begin{table}[t]
\caption{\textbf{$\alpha$ selection for ITOP dataset (top view).} $p$-values and average validation MSE per $\alpha$.}
\label{tab:statsAL1iTOPtview}
\scriptsize 
\begin{center}
\setlength{\tabcolsep}{0.2em} %
\begin{tabular}{|l|c c c c c c c c |}
\hline
 $\alpha$ &\emph{0}& \emph{0.01} & \emph{0.02}& \emph{0.04}& \emph{0.05}& \emph{0.06}& \emph{0.08}& \emph{0.1}\\
  \hline \emph{0} & - & 0.5476 & 1.0000 & 0.6905 & 1.0000 & 1.0000 & \textbf{0.8413} & 0.8413  \\
  \hline \emph{0.01} & 0.5476 & - & 0.5476 & 1.0000 & 0.5476 & 0.3095 & \textbf{0.1508} & 0.5476  \\
  \hline \emph{0.02} & 1.0000 & 0.5476 & - & 0.6905 & 1.0000 & 1.0000 & \textbf{0.8413} & 1.0000  \\
  \hline \emph{0.04} & 0.6905 & 1.0000 & 0.6905 & - & 0.8413 & 0.6905 & \textbf{0.6905} & 0.5476  \\
  \hline \emph{0.05} & 1.0000 & 0.5476 & 1.0000 & 0.8413 & - & 1.0000 & \textbf{0.6905} & 0.8413  \\
  \hline \emph{0.06} & 1.0000 & 0.3095 & 1.0000 & 0.6905 & 1.0000 & - & \textbf{0.6905} & 1.0000  \\
  \hline \emph{0.08} & 0.8413 & 0.1508 & 0.8413 & 0.6905 & 0.6905 & 0.6905 & - & 0.6905  \\
  \hline \emph{0.1} & 0.8413 & 0.5476 & 1.0000 & 0.5476 & 0.8413 & 1.0000 & \textbf{0.6905} & -  \\
\hline 
\hline
\textit{MSE} & 0.9865  &  1.0144    &0.9869  &  1.0013  &  0.9948 &   0.9918&    \textbf{0.9767}&    0.9861\\
\hline
\end{tabular} 
\end{center}
\end{table}

\section*{Acknowledgements}
This project has been funded under projects TIN2016-75279-P and IFI16/00033 (ISCIII) of Spain Ministry of Economy, Industry and Competitiveness, and FEDER.
Thanks to NVidia for donating the GPU Titan Xp used for the experiments presented in this work.
We also thank Shafaei and Little for providing their error and precision results used in our comparative plots.

\clearpage


\bibliographystyle{model2-names}

\bibliography{longstrings,bibAVA,local}

\end{document}